\pdfoutput=1
\documentclass[11pt]{article}

\usepackage[margin=1in]{geometry}

\usepackage[utf8]{inputenc} 
\usepackage[T1]{fontenc}    
\usepackage[backref,colorlinks,citecolor=blue,linkcolor=magenta,bookmarks=true]{hyperref}       
\usepackage{url}            
\usepackage{booktabs}       
\usepackage{amsfonts}       
\usepackage{nicefrac}       
\usepackage{microtype}      
\usepackage{xcolor}         
\usepackage{amsfonts}  
\usepackage{amsthm}
\usepackage{amsmath}
\usepackage{amssymb}
\usepackage{mathtools}
\usepackage{mathrsfs}
\usepackage{algorithm}
\usepackage{algpseudocode}
\usepackage[nameinlink]{cleveref}
\usepackage{graphicx}
\usepackage[normalem]{ulem}
\usepackage{titletoc}
\usepackage[numbers]{natbib}
\usepackage{parskip}

\usepackage{color-edits}
\addauthor{mq}{blue}
\addauthor{kw}{orange}

\newcommand{\Bern}{\mathsf{Bern}}

\newcommand{\calB}{\mathcal{B}}
\newcommand{\calE}{\mathcal{E}}

\newcommand{\calO}{\mathcal{O}}
\newcommand{\calT}{\mathcal{T}}
\newcommand{\Egood}{\calE^{\mathrm{good}}}
\newcommand{\Ex}[2]{\operatorname*{\mathbb{E}}_{#1}\left[#2\right]} 
\newcommand{\guess}{\mathsf{guess}}
\newcommand{\Lhat}{\widehat{L}}
\newcommand{\LHS}{\mathrm{LHS}}
\newcommand{\pr}[2]{\Pr_{#1}\left[#2\right]}
\newcommand{\Reg}{\mathrm{Reg}} 
\newcommand{\remain}{\mathsf{remain}}
\newcommand{\RHS}{\mathrm{RHS}}
\newcommand{\zo}{\{0, 1\}}

\newtheorem{definition}{Definition}
\newtheorem{lemma}{Lemma}
\newtheorem{proposition}{Proposition}
\newtheorem{theorem}{Theorem}
\newtheorem{remark}{Remark}
\newtheorem{corollary}{Corollary}
\newtheorem{example}{Example}

\usepackage[utf8]{inputenc} 
\usepackage[T1]{fontenc}    
\usepackage{hyperref}       
\usepackage{url}            
\usepackage{booktabs}       
\usepackage{amsfonts}       
\usepackage{nicefrac}       
\usepackage{microtype}      
\usepackage{xcolor}         

\title{Adversarially Batched Online Learning}

\author{
Mingda Qiao and Kerui Wu\\[0.7em]
University of Massachusetts Amherst\\[0.15em]
\texttt{\{mqiao,keruiwu\}@umass.edu}
}

\date{}

\begin{document}

\maketitle


\begin{abstract}\noindent
We introduce \emph{adversarially batched online learning}, where the learner may update its decision rule only at a restricted set of time steps. Equivalently, the time horizon is partitioned into \emph{batches}, within which the learner must use a fixed policy. We study this model for four canonical problems: stochastic bandits, online convex optimization, prediction with expert advice, and adversarial bandits. For stochastic bandits, we identify a universal complexity measure based on relative batch growth and establish both gap-dependent and gap-independent regret bounds.
We show that the multiplicative penalty relative to the standard gap-dependent complexity can grow superlinearly in batch hardness.
For the remaining three, we show that the optimal regret is tightly characterized by the $2$-norm of the batch-length sequence. The upper bound is achieved by a \emph{batch-oblivious} learner who does not know the batch structure, via a generic adaptive doubling trick tailored to this complexity measure. In contrast, the lower bound holds even for \emph{batch-aware} learners.
\end{abstract}

\section{Introduction}
Online learning provides a framework for sequential decision-making in uncertain environments. In standard settings, the learner plays an action at each round and observes the associated losses, which may immediately factor into the decision in the next round. The objective is to minimize the cumulative regret, defined as the difference between the total loss of the learner and that of the best fixed action in hindsight.
While this framework has been studied extensively (e.g., see \cite{cesa2006prediction,LS20}), the assumption that the learner can freely revise its policy after every round is often violated in practice. Frequent policy updates may be infeasible due to physical, logistical, or system constraints: outcomes may only become available after a trial phase, human annotations may need to be collected and validated together, and system-level updates may require synchronization.

This work addresses \emph{Batched Online Learning}, where the learner is only allowed to update their decision rules at a set of specific time points. Prior work in this domain has largely focused on scenarios where the learner designs the batch structure to optimize the trade-off between the regret and the number of batches (e.g.,~\cite{gao2019batched,esfandiari2021regret}). 
In contrast,
we investigate the \emph{Adversarially Batched} setting, where the update points are imposed by an external environment (or adversary) and are immutable from the learner's perspective. For example, a clinical trial may release cumulative outcome data only at irregular interim review dates, a crowdsourcing platform may return validated annotations only after unpredictable quality-control cycles, and a distributed system may synchronize logs or deploy updates only when communication windows become available at unpredictable times.

The core challenge in our setting is that the learner must make a commitment at the beginning of each batch that cannot be revised until the batch ends, potentially accruing significant regret if the batch is long. We study this setting under four classic online learning problems: \emph{Stochastic Bandits}, \emph{Online Convex Optimization (OCO)}, \emph{Experts Problem}, and \emph{Adversarial Bandits}. 


The main contributions of this work are summarized below. We formally introduce our setup in \Cref{sec:setup} and state our results in \Cref{sec:results}.
\begin{itemize}
    \item We formulate a model of adversarially batched online learning, where the environment specifies the time steps on which the learner is allowed to update its decision rule.

    \item For stochastic bandits, we identify a universal batch-complexity measure that captures the multiplicative regret increase over the non-batched setting. We prove both gap-dependent and gap-independent regret bounds that depend on this quantity.
    
    \item For the other three problems, we show that the optimal regret is characterized by the $2$-norm of the batch length sequence by proving nearly matching upper and lower bounds. The upper bound is achieved by a \emph{batch-oblivious} learner that does not know the batch structure, while the lower bound applies to \emph{batch-aware} learners as well. 

\end{itemize}

\subsection{Problem Setup}\label{sec:setup}
In standard online learning settings, the learner interacts with the environment over a sequence of time steps $t \in [T] \coloneqq \{1,2,\ldots,T\}$. At each time step, the learner makes a decision based on the feedback observed from previous rounds, receives feedback from the environment, and updates its decision rule. The goal of the learner is to minimize the cumulative regret, which quantifies the sub-optimality of its actions compared to the optimal fixed action in hindsight.

\noindent\textbf{Our model: Adversarially Batched Online Learning.}~
Departing from the standard model,
we study an \emph{adversarially batched} setting: 
the learner will receive the cumulative loss at certain update time points, where both the loss at each round and the update points are determined by the environment (adversarially). In other words,
the environment segments the time horizon into $B$ batches, and the learner can only update its strategy at the end of each batch. We denote the set of update time steps by $\calT = \{t_1, t_2, \dots, t_B\}$. Equivalently, the instance can be described by its \emph{batch structure} $\calB = (b_1, b_2, \dots, b_B) = (t_1 - 0, t_2 - t_1, \dots, t_B - t_{B-1})$.\footnote{We view the end of the horizon as an update point, i.e., $t_B = T$.}
We will primarily use the batch structure $\calB$ to specify an instance. We also write $L_i^{(j)} = \sum_{k=t_{j-1}+1}^{t_j} l_i^{(k)}$ as the cumulative loss of action $i \in [N]$ within batch $j \in [B]$.

We further distinguish between two settings regarding the learner's knowledge of the batch structure. In the \emph{batch-aware} setting, the learner is given the entire update time set $\calT$ (or equivalently the batch structure $\calB$) before the learning process begins. The learner may then leverage this information, e.g., to tune the learning rate. In contrast, in the \emph{batch-oblivious} setting, the learner does not know the future update times in advance. Instead, the learner only gets notified that a batch has ended at each time step $t_j \in \calT$, and still does not know the future update times $t_{j+1}, \ldots, t_B$. 

We instantiate the model of adversarially batched online learning in four canonical settings: \emph{stochastic bandits}, \emph{online convex optimization (OCO)}, \emph{experts problem}, and \emph{adversarial bandits}.

\noindent\textbf{Stochastic Bandits.}~
In \textit{stochastic} Multi-Armed Bandits (MAB), each of the $N$ arms is associated with a fixed but unknown reward distribution over $[0, 1]$. Each pull of an arm results in a reward independently sampled from that distribution. In the adversarially batched setting, the learner commits to a policy that specifies the sequence of arms to be pulled (e.g., a round-robin policy that cycles through $1, 2, \ldots, N$). The learner can only update this policy at the end of each batch. Let $\mu_i$ denote the mean reward of arm $i \in [N]$, and suppose that the learner pulls arm $i_t \in [N]$ at time $t$. Then, the learner incurs a regret of
$
\Reg_N^{(T)} \coloneqq T \cdot \max_{i^* \in [N]} \mu_{i^*} - \sum_{t=1}^{T} \mu_{i_t}.
$

\noindent\textbf{OCO.}~
Let $\mathcal{K} \subseteq \mathbb{R}^d$ be a nonempty compact convex action set and $D := \sup_{x, y \in \mathcal{K}} \| x - y \|_2$ be its Euclidean diameter. At every round $t \in [T]$, the environment chooses a convex loss function $f_t : \mathcal{K} \rightarrow \mathbb{R}$. We assume that $f_t$ is $G$-Lipschitz over $\mathcal{K}$, i.e., every subgradient $g \in \partial f_t(x)$ satisfies $\| g \|_2 \leq G$. In the adversarially batched setting, the learner must use the same action $x^{(j)} \in \mathcal{K}$ throughout each batch $j$. At the end of each batch, the learner receives first-order feedback for the losses incurred in that batch. Let $z_j = \sum_{t=t_{j-1}+1}^{t_j} g_t$ denote the aggregate subgradient in batch $j$ where $g_t \in \partial f_t(x^{(j)})$. Since each $f_t$ is $G$-Lipschitz, 
$
\| z_j \|_2 \leq G b_j.
$
The regret is measured against the best fixed action in hindsight:
$
\mathrm{Reg} = \sum_{j=1}^B \sum_{t=t_{j-1}+1}^{t_j} f_t (x^{(j)}) - \min_{x^* \in \mathcal{K}} \sum_{t=1}^T f_t(x^*).
$

\noindent\textbf{Experts Problem.}~
In the \emph{experts problem}, the learner interacts with $N$ experts. At every round, the learner selects a distribution over the experts and incurs the expected loss under this distribution. The losses of all experts are then revealed to the learner. Formally, let $p_i^{(j)}$ denote the probability assigned to expert $i \in [N]$ at batch $j \in [B]$.\footnote{The learner must select the same probability distribution within each batch.} The regret of the learner in the adversarially batched experts problem is then defined as the difference between the cumulative loss of the learner and that of the best fixed expert in hindsight:
$
    \Reg_N^{(B)} = \sum_{j=1}^{B} \sum_{i=1}^N \left(p_i^{(j)}L_i^{(j)} \right) -\min_{i^* \in [N]} \sum_{j=1}^{B} L_{i^*}^{(j)}.
$
\noindent\textbf{Adversarial Bandits.}~
Similar to the experts problem, an adversarial MAB instance is specified by the losses $l_i^{(t)} \in [0, 1]$ of all actions (or ``arms'') $i \in [N]$ at all time steps $t \in [T]$. The difference is that the learner, instead of observing all losses after each round, only sees the loss of the chosen arm. Specifically, we study the stricter setting in which the learner commits to a single arm for the entire batch, and only learns the losses associated with that action throughout the batch. 



We measure performance by the expected static regret, which is also known as the \emph{pseudo-regret}, which compares the learner's expected cumulative loss with the lowest expected loss over all arms in hindsight, i.e.,
$
\overline{\Reg}_N^{(B)} = \mathbb{E} \left[ \sum_{j=1}^{B} \sum_{i=1}^{N} p_i^{(j)} L_i^{(j)} - \min_{i^* \in [N]}\sum_{j=1}^{B} L_{i^*}^{(j)} \right].
$
All expectations in the upper bounds are over the learner's internal randomization, and over the adversary's initial randomization if it draws the oblivious loss table from a distribution. 


\subsection{Our Results}\label{sec:results}
For each problem, we characterize the effect of the batch structure through matching regret bounds governed by different complexity measures.

\subsubsection{Stochastic Bandits}
\paragraph{Upper Bounds.}
Let $\Delta_i \coloneqq \max_{i^* \in [N]} \mu_{i^*} - \mu_i \geq 0$ denote the \emph{gap} of arm $i$. The well-known Upper Confidence Bound (UCB) algorithm~\cite{auer2002nonstochastic} has a gap-dependent regret bound of $\calO\left( (\log T) \cdot \sum_{i: \Delta_i > 0} 1/\Delta_i \right)$ and a gap-independent bound of $\calO(\sqrt{NT \log N})$. 

For the batched setting, \cite{gao2019batched} studied an easier setting where the update points are freely chosen by the learner. They proposed an algorithm and proved regret upper bounds under \emph{specific} batch structures. In particular, given the time horizon $T$ and the batch number $B$, the batch structure $b_j = \Theta(T^{j/B})$ leads to an expected regret upper bound of $\calO \left( T^{1/B} \log N \log(NT) \frac{N}{\Delta_{\min}} \right)$, where $\Delta_{\min} := \min_{i: \Delta_i > 0} \Delta_i$ is the minimum non-zero gap.

For our adversarially batched setting, we first introduce a universal complexity measure $\alpha$ that captures the hardness of an \emph{arbitrary} batch structure in stochastic bandits. 

\begin{definition}\label{def:hardness-metric}
The hardness metric $\alpha$ of a batch structure $\calB = (b_1, b_2, \dots, b_B)$ is 
\[
\alpha \coloneqq \max_j \frac{b_j}{\sum_{k=1}^{j-1} b_k} = \max_j \frac{b_j}{t_{j-1}},
\]
where, for brevity, we define $t_0 = 1$ to avoid division by zero.\footnote{Equivalently, we can write $\alpha = \max \left\{ b_1, \max_{j \geq 2} \frac{b_j}{t_{j-1}} \right\}$ without the convention that $t_0 = 1$.}
\end{definition}

Intuitively, $\alpha$ captures the largest mismatch between a batch length and the evidence available before it. In batch $j$, the algorithm commits to $b_j$ rounds using only the previous $t_{j-1}$ observations; hence a large ratio $b_j/t_{j-1}$ means long blind commitments given limited data, while small ratios mean information keeps pace with commitment length.

\begin{remark}
\label{remark:stochastic_alpha_upper}
Under the batch structure $b_j = \Theta \left( T^{j/B} \right)$, we have $\alpha = \Theta \left( T^{1/B} \right)$, so the regret bound of~\cite{gao2019batched} can be simplified to $\calO \left( \alpha \log N \log(NT) \frac{N}{\Delta_{\min}} \right) = \widetilde{\calO}(\alpha N / \Delta_{\min})$. 
\end{remark}

As a warm-up, we extend the $\alpha$-dependent regret bound of~\cite{gao2019batched} (\Cref{remark:stochastic_alpha_upper}) to all batch structures: in \Cref{propo:existing_alpha_bound} (\Cref{sec:stochastic_alpha_bound}), we prove an $\widetilde{\calO}(\min\left\{ T, \alpha N/\Delta_{\min} \right\})$ regret bound for any batch structure. Our main technical contribution then goes further: we propose a successive-elimination algorithm with merging operations in \Cref{sec:stochastic}, and prove both gap-dependent and gap-independent regret guarantees for adversarially batched stochastic bandits.

\begin{theorem}
\label{thm:stochastic_bandit}
For adversarially batched stochastic bandits with $N$ arms and an arbitrary batch structure $\calB$ with hardness metric $\alpha$, there is an algorithm with a gap-dependent expected regret of
$
\calO\left( \min\left\{ T, \alpha^2 \log(NT) \sum_{i: \Delta_i > 0} \frac{1}{\Delta_i} \right\}\right).
$
\end{theorem}
\begin{theorem}
\label{thm:stochastic_gap_ind}
For adversarially batched stochastic bandits with $N$ arms and an arbitrary batch structure $\calB$ with hardness metric $\alpha$, there is an algorithm with a gap-independent expected regret of
$
\widetilde{\calO}\left( \min \left\{T,\sqrt{\alpha TN} \right\}\right).
$
\end{theorem}
Compared with \cite{gao2019batched}, our results apply to any batch structure $\calB$, rather than only to specially designed ones. Our gap-dependent bound (\Cref{thm:stochastic_bandit}) replaces the coarse worst-arm dependence $N/\Delta_{\min}$ by the standard UCB-type complexity $\sum_{i:\Delta_i>0}\Delta_i^{-1}$, yielding an improvement of up to a factor of $N$ in its dependence on the gaps. This refinement costs an additional factor of $\alpha$, but only for the gap-dependent bound. The gap-independent bound, on the other hand, has a better (square root) dependence on the batch hardness measure $\alpha$. In particular, when no batch is significantly longer than the cumulative length of the preceding batches, i.e., $\alpha = \mathcal{O}(1)$, both bounds recover the standard non-batched regret rates up to polylogarithmic factors.

\paragraph{Lower Bounds.}
On the lower-bound side, we show that a polynomial dependence on $\alpha$ is
unavoidable. Firstly, in \Cref{app:existing_stochastic_lower}, we extend the analysis of \cite{gao2019batched} to establish both gap-dependent and gap-independent worst-case lower bounds. 


\begin{proposition}
\label{thm:existing_stochastic_lower}
Suppose $t_{j^* - 1} \geq CN$, where $j^* \in \arg\max_j b_j/t_{j-1}$ and $C$ is a sufficiently large constant. Then, there is a hard instance of adversarially batched stochastic bandits such that every algorithm suffers a gap-dependent regret lower bound of $\Omega\!\left( \alpha \sum_{i: \Delta_i > 0} \Delta_i^{-1} \right)$. 
\end{proposition}

\begin{theorem}
\label{thm:existing_stochastic_gap_independent_lower}
Suppose $t_{j^* - 1} \geq C_1N$ and $t_{j^* - 1} \geq C_2\frac{T}{\alpha}$, where $j^* \in \arg\max_j b_j/t_{j-1}$ and $C_1, C_2$ are sufficiently large constants. Then, there is a hard instance of adversarially batched stochastic bandits such that every algorithm suffers a gap-independent regret lower bound of $\Omega\!\left( \sqrt{\alpha N T} \right)$.
\end{theorem}

\Cref{thm:existing_stochastic_gap_independent_lower} gives a gap-independent lower bound that matches the upper bound in \Cref{thm:stochastic_gap_ind} up to logarithmic factors on a suitable family of hard batch structures. On the other hand, for the gap-dependent setting, \Cref{thm:stochastic_bandit} and \Cref{thm:existing_stochastic_lower} still leave a multiplicative gap of $\alpha$. To narrow this gap, in \Cref{sec:stochastic_gap_dependent_lower} we construct a carefully designed hard instance and establish a stronger gap-dependent lower bound with an $\alpha^{3/2}$ dependence.

\begin{theorem}
\label{thm:stochastic_worst_case_lower}
There exists a hard instance of adversarially batched stochastic bandits such that every algorithm suffers an expected cumulative regret of 
$
\Omega\!\left( \alpha^{\frac{3}{2}} \sum_{i:\Delta_i > 0} \frac{1}{\Delta_i} \right).
$
\end{theorem}

Moreover, we show in \Cref{sec:stochastic_algorithm_lower} that our gap-dependent analysis necessarily leads to an $\Omega(\alpha^{7/4})$ factor in the regret bound.



\subsubsection{Convex Optimization-based Problems}
\paragraph{Batch-Aware Setting.}
In \Cref{app:oco_upper,sec:expert_upper,app:adversarial_bandit_upper}, we prove the following regret upper bounds for the other three problems with convex optimization formulations under the adversarially batched model, including OCO, the experts problem (which can be formulated as an Online Linear Optimization (OLO) over the probability simplex), and the adversarial bandits (which can be formulated as an instance of bandit convex optimization). The bounds are obtained by classic algorithms: Online Gradient Descent (OGD)~\citep{zinkevich2003online}, Hedge~\citep{freund1997decision, cesa2006prediction}, and EXP3~\citep{auer2002nonstochastic}. 

\begin{theorem}
\label{thm:oco_expert_adversarial_upper}
For any batch structure $\mathcal{B}$, the following regret guarantees hold:
\begin{enumerate}
    \item In the adversarially batched OCO problem, projected OGD with learning rate $\eta>0$ satisfies $\mathrm{Reg} \leq \frac{D^2}{2 \eta} + \frac{\eta G^2}{2} \| b \|_2^2$. Setting $\eta = \frac{D}{G \| b \|_2}$ gives a regret upper bound of $\mathcal{O}(DG \| b \|_2)$.
    \item In the adversarially batched experts problem with $N$ experts, Hedge with learning rate $\eta>0$ satisfies $\mathrm{Reg} \leq \frac{\ln N}{\eta} + \frac{\eta}{8}\|b\|_2^2$. Setting $\eta = \frac{\sqrt{8 \ln N}}{\|b\|_2}$ gives a regret upper bound of $\calO\left(\min\left\{\|b\|_2 \sqrt{\log N}, T\right\}\right)$.
    \item In the adversarially batched adversarial bandits problem with $N$ arms, EXP3 with learning rate $\eta > 0$ satisfies $\overline{\mathrm{Reg}} \leq \frac{\ln N}{\eta} + \frac{\eta N}{2}\|b\|_2^2$. Setting $\eta = \frac{\sqrt{2 \ln N}}{\|b\|_2 \sqrt{N}}$ bounds the pseudo-regret by $\calO\left( \min \left\{\|b\|_2\sqrt{N \log N}, T \right\} \right)$.
\end{enumerate}
\end{theorem}


We complement the upper bounds with matching or nearly-matching worst-case lower bounds (up to logarithmic factors) in \Cref{app:oco_lower,sec:expert_lower,app:adversarial_bandit_lower}, demonstrating that the dependence on $\|b\|_2$ is unavoidable. For OCO, the lower bound holds for every fixed batch structure, whereas for experts and adversarial bandits we establish the corresponding worst-case lower bounds under an additional mild assumption on the maximum batch length.

\begin{theorem}
\label{thm:oco_expert_adversarial_lower}
The following lower bounds hold for adversarially batched online learning:
\begin{enumerate}
    \item For every batch structure $\mathcal{B}$, every algorithm for adversarially batched OCO suffers an expected regret of $\Omega\left( DG\|b\|_2 \right)$.
    \item For every $N \geq 2$ and batch structure $\mathcal{B}$, every algorithm has a worst case regret of $\Omega(\|b\|_2)$. Moreover, assuming $\max_{i \in [B]} b_i \leq \mathcal{O}(\|b\|_2 / \sqrt{\log N})$, the regret is at least $\Omega\left(\|b\|_2 \sqrt{\log N} \right)$. 
    \item For every $N \geq 2$ and batch structure $\mathcal{B}$, every algorithm has a worst-case pseudo-regret of $\Omega\left(\min\left\{ \|b\|_2\sqrt{N}, \frac{\|b\|_2^2}{b_{\max}} \right\}\right)$, where $b_{\max} = \max_{j \in [B]} b_j$. In particular, if $b_{\max} \leq \mathcal{O}(\|b\|_2 \sqrt{N})$, then the pseudo-regret is lower bounded by $\Omega\left( \| b \|_2 \sqrt{N} \right)$. 
\end{enumerate}
\end{theorem}

\paragraph{Batch-Oblivious Setting.}
The algorithms in \Cref{thm:oco_expert_adversarial_upper} are batch-aware, since the optimal learning rate depends on the quantity $\|b\|_2$. Towards extending the upper bound to the batch-oblivious setting, we show in \Cref{app:scale_free_reduction,app:delayed_feedback_reduction} that algorithms from scale-free and delayed-feedback online learning can be adapted. 
Specifically, for scale-free online learning, we treat each batch as a meta-round and deploy SOLO FTRL \citep{orabona2018scale} for OCO and the experts problem, and UMAB-NN \citep{chen2023improved} for adversarial bandits, to recover the same order of upper bounds. For delayed-feedback models, we view batched feedback as a special case of delayed feedback and apply the delay-adaptive algorithms of \cite{joulani2016delay} to OCO and the experts problem. We note that existing delayed-feedback results do not directly apply to the stricter adversarial bandits setting studied here, as they typically allow the learner to select new arms while feedback from previous rounds remains outstanding. 


\subsubsection{An Adaptive Doubling Trick}
The reductions above recover batch-oblivious guarantees through existing scale-free or delay-adaptive algorithms. We additionally provide a generic conversion that preserves the choice of the base algorithm: As shown in \Cref{thm:adaptive_doubling}, any restartable algorithm satisfying the fixed-learning-rate bound and instantaneous-regret condition, which bounds the regret incurred in any single round against any comparator by $\mathcal{O}(\sqrt{AD})$, can be made batch-oblivious using only learning-rate selection and restarts, with no additional logarithmic factor in its regret guarantee.


\begin{theorem}
\label{thm:adaptive_doubling}
Consider a base online learning algorithm $\mathcal{A}_\eta$, parametrized by a learning rate $\eta > 0$. Suppose that there exist known quantities $A, D > 0$, such that whenever $\mathcal{A}_\eta$ is run from scratch on any batch structure $\mathcal{B}$, its regret satisfies $\mathrm{Reg}(\mathcal{A}_\eta) = \mathcal{O} \left( \frac{A}{\eta} + D \eta \|b\|_2^2 \right)$. Suppose the instantaneous regret against any comparator is at most $L$, where $L=\calO(\sqrt{AD})$. Then, there exists a batch-oblivious algorithm whose regret satisfies $\mathcal{O} \left( \sqrt{AD} \|b\|_2 \right)$.
\end{theorem}
In particular, \Cref{thm:adaptive_doubling} allows OGD, Hedge, and EXP3 to retain their batch-aware rates without requiring advance knowledge of the batch structure. The main difficulty is that a batch may overshoot the current budget by an arbitrarily large amount before its length becomes known. \Cref{propo:failure_doubling_trick} (proved in \Cref{sec:naive_doubling_proof}) shows a na\"ive implementation of the standard doubling trick~\citep{cesa2006prediction,besson2018doubling} can fail dramatically. In contrast, our analysis controls this batch separately, and the next budget is calibrated to the realized sum of squared batch lengths. 
\begin{proposition}
\label{propo:failure_doubling_trick}
    There exists a family of instances of the adversarially batched experts problem on which the na\"ive doubling-trick adaptation of Hedge incurs an $\Omega(T)$ regret, whereas a batch-aware algorithm can achieve an $\calO(T^{3/4})$ regret.
\end{proposition}

\subsection{Related Work}
\label{sec:related_work}
Our work builds on the vast literature on online learning, including stochastic MAB~\citep{lai1985asymptotically,auer2002finite}, OCO~\citep{zinkevich2003online}, the experts problem~\citep{littlestone1994weighted,freund1997decision}, and adversarial bandits~\citep{auer2002nonstochastic}. We refer the reader to~\cite{cesa2006prediction,LS20} for a comprehensive introduction.

\noindent\textbf{Selective Prediction with Limited Selectivity.}~
Most similar to our setup is the recent work of~\cite{LQ25}, which builds on the \emph{selective prediction} formulation of \cite{Drucker13,QV19} and studies a variant in which the forecaster can only initiate predictions at a prescribed subset of time steps. Their setting can be viewed as prediction-focused in the sense that restrictions are imposed on when predictions may be made, whereas we generalize this timing constraint to regret minimization in online learning, covering the stochastic bandits, OCO, the experts problem, and adversarial bandits.

\noindent\textbf{Batched Bandits.}~
The constraint of batched feedback in multi-armed bandits has garnered significant attention recently. The seminal work of \cite{perchet2016batched} characterized the minimax regret for stochastic bandits with a limited number of batches, showing that $\mathcal{O}(\log T)$ batches are sufficient to achieve the optimal $\mathcal{O}(\sqrt{T})$ regret. \cite{gao2019batched} extended this by analyzing the trade-off between the number of batches and regret, introducing a ``grid'' structure. \cite{esfandiari2021regret} further improved these bounds for specific batch structures. \cite{jin2021almost} studied the anytime version of batched bandits and developed algorithms that allow the learner to adaptively choose update points while approaching the minimal batch complexity. Crucially, these works generally assume the learner can design the grid (e.g., the batch structure) to minimize the regret. Our setting differs fundamentally: the batch structure is \emph{adversarial} and imposed by the environment, forcing the learner to act under external constraints rather than optimizing the schedule.

\noindent\textbf{Scale-Free and Loss-Adaptive Online Learning.}~
Existing online learning frameworks can potentially be deployed in the adversarially batched model through batch-level aggregation. In this study, we show batch-oblivious reductions in the experts problem and adversarial bandits using scale-free and loss-adaptive online learning. Specifically, \cite{orabona2018scale} developed scale-free online learning algorithms whose regret guarantees adapt to the observed magnitudes of the loss vectors. By treating each batch as a single meta-round, SOLO FTRL can be directly deployed for OCO and the experts problem while retaining the desired dependence on the realized batch lengths. For adversarial bandits, an analogous reduction uses the unbounded-loss bandit algorithm of \cite{chen2023improved}. 

\noindent\textbf{Delayed Feedback Model.}~
Our problem is also related to online learning with delayed feedback~\citep{mesterharm2005line,joulani2013online,joulani2016delay,schlisselberg2026improved}. For \emph{OCO} and the \emph{experts problem}, similar to scale-free reductions (\Cref{app:scale_free_reduction}), we show in \Cref{app:delayed_feedback_reduction} that delayed-feedback models can be reduced to obtain the same order of batch-oblivious regret upper bounds. In contrast, existing delayed-bandit results do not directly apply to our \emph{adversarial bandits} problem, which requires the learner to commit to a single arm throughout each batch. For \emph{stochastic bandits}, existing delayed-feedback results \citep{joulani2013online,schlisselberg2026improved} typically incur an additive penalty on top of the standard regret, controlled by \emph{absolute delay}. In the adversarially batched setting, this can be expressed as $\mathcal{O}\left( \min\left\{ T, \sum_{i:\Delta_i>0}\frac{\log T}{\Delta_i} + b_{\max}\sum_{i:\Delta_i>0}\Delta_i \right\} \right)$ or $\widetilde{\mathcal{O}}\left( \min\left\{ T, \sqrt{NT}+N b_{\max} \right\}\right)$, where $b_{\max}:=\max_j b_j$. In contrast, our analysis provides a multiplicative characterization through the \emph{relative growth} measure $\alpha$. Consequently, we provide complementary bounds that can be substantially sharper whenever batch sizes are large but remain controlled relative to the accumulated history. For example, under geometrically growing batches, the maximum batch length and the number of outstanding observations can be $\Theta(T)$, whereas $\alpha=\Theta(1)$.

\section{Stochastic Bandits}\label{sec:stochastic}
Our algorithm (\Cref{algo:stochastic_bandits}) builds on the Batched Successive Elimination algorithm of~\cite{gao2019batched}, with an additional \emph{merging} operation.
Specifically, let $\gamma > 0$ be a constant parameter associated with the UCB bound. Let $A_j$ denote the set of active arms at the beginning of batch~$j$ and $n_j \coloneqq |A_j|$. Then, $S_j \coloneqq \sum_{k=1}^j \frac{b_k}{n_k}$ is the total number of times arm $i \in A_j$ is pulled by the end of batch~$j$ (i.e., at time step $t_j$)\footnote{Under the least-pulled-arm rule, the cumulative pull counts of active arms differ by at most one.}. \Cref{algo:stochastic_bandits} eliminates every arm with an empirical mean that is lower, by a margin of at least $\sqrt{\frac{\gamma \ln(NT)}{S_j}}$, than the highest empirical mean $\hat{\mu}_{a^*}$.

\noindent\textbf{A Gap-Dependent Upper Bound.}~
To analyze \Cref{algo:stochastic_bandits}, we define $j_i$ to be the index of the block after which arm $i$ is \emph{intended} to be eliminated. Recall that $\Delta_i = \max_{i^* \in [N]} \mu_{i^*} - \mu_i$ is the gap of arm~$i$. Then, $j_i$ is formally defined as $j_i \coloneqq \min\left\{j \in [B]: S_j \ge \frac{4 \gamma \ln(NT)}{\Delta_i^2}\right\}$ if this set is nonempty; otherwise, we define $j_i = B + 1$.

We will condition on a ``good event'' $\Egood$ defined as follows: (1) the optimal arm $i^*$ remains active throughout the algorithm; (2) every sub-optimal arm $i$ with $\Delta_i>0$ and $j_i \le B$ is eliminated by the end of batch $j_i$.  \Cref{lemma:stochastic_est_fail}, proved in \Cref{sec:stochastic_est_fail}, upper bounds the probability that $\Egood$ does not happen by $1/T$. The contribution of this case to the expected regret is at most $T \cdot \frac{1}{T} = 1$. Then, \Cref{lemma:stochastic_cond_regret} upper bounds the regret assuming event $\Egood$. These two lemmas together yield \Cref{thm:stochastic_bandit}. We provide a brief proof sketch below for \Cref{lemma:stochastic_cond_regret} and defer the full proof to \Cref{app:stochastic_cond_regret}.


\begin{algorithm*}[t]
\caption{Batched Arm Elimination with Merging}
\label{algo:stochastic_bandits}
\begin{algorithmic}[1]
\Require Time horizon $T$. Number of arms $N$. Parameter $\gamma > 0$.
\State Initialize $j \gets 1$; $A_1 \gets [N]$; $b_j \gets 0$; \Comment{all arms are active}
\For{$t = 1, 2, \ldots, T$}
    \State Pull the least-pulled arm in $A_j$; \Comment{pull active arms in a round-robin fashion}
    \State $b_j \gets b_j + 1$; \Comment{track batch length}
    \If{$t$ is among the updating time steps}
        \State Update the empirical mean $\hat{\mu}_a$ for each arm $a \in A_j$;
        \If{$j=1 \;\mathrm{or}\; b_j \geq 2 \cdot b_{j-1}$} \Comment{the merged batch is long enough}
            \State $A_{j+1} \gets \left\{a \in A_j: \max_{a^* \in A_j} \hat{\mu}_{a^*} - \hat{\mu}_a \leq \sqrt{\frac{\gamma \log( NT)}{S_j}}\right\}$; \Comment{eliminate arms}
            \State $j \gets j + 1$; $b_j \gets 0$;
        \EndIf 
    \EndIf
\EndFor
\end{algorithmic}
\end{algorithm*}

\begin{lemma}
\label{lemma:stochastic_est_fail}
    $\Pr[\Egood] \ge 1 - 1/T$.
\end{lemma}

\begin{lemma}
\label{lemma:stochastic_cond_regret}
Conditioned on $\Egood$, \Cref{algo:stochastic_bandits} has regret
$
\calO\left(\alpha^2 \log(NT) \sum_{i: \Delta_i > 0} \Delta_i^{-1}\right).
$
\end{lemma}

\begin{proof}[Proof sketch of \Cref{lemma:stochastic_cond_regret}]
Suppose that a sub-optimal arm $i$ remains active at the beginning of block $j$, i.e., it has not been eliminated by the end of block $j-1$. By the definition of $j_i$ and the good event, we have $\Delta_i \le \sqrt{\frac{4\gamma \ln(NT)}{S_{j-1}}}$. Then, the total regret incurred by arms eliminated at the end of block $j$ is at most
$
    (n_j-n_{j+1}) S_j
    \sqrt{\frac{4\gamma \ln(NT)}{S_{j-1}}},
$
where $n_j$ is the number of active arms during block $j$ and $S_j = \sum_{k=1}^j \frac{b_k}{n_k}$ is the number of pulls per active arm up to the end of block $j$. Assume for simplicity that the block lengths grow geometrically except for the last one, i.e.,
$b_j \ge 2b_{j-1}$ for all $j \in [B - 1]$ while $b_B \le 2b_{B-1}$. Under this assumption, we can show that $t_j < 2b_j$ and $S_j=\Theta\!\left(b_j / n_j\right)$. The total conditional regret is then bounded by $\calO(\sqrt{\log(NT)})$ times
\begin{equation}\label{eq:stochastic-bandits-LHS}
    \sum_{j=1}^{B}
    (n_j-n_{j+1})
    \frac{b_j}{n_j}
    \sqrt{\frac{n_{j-1}}{b_{j-1}}}.
\end{equation}

On the other hand, the $\sum_{i:\Delta_i>0}1/\Delta_i$ term in the target bound is at least, up to a $\sqrt{\log(NT)}$ factor,
\begin{equation}\label{eq:stochastic-bandits-RHS}
    \sum_{j=1}^{B}
    (n_j-n_{j+1})
    \sqrt{\frac{b_{j-1}}{n_{j-1}}}.
\end{equation}
To bound \Cref{eq:stochastic-bandits-LHS} by \Cref{eq:stochastic-bandits-RHS}, we split blocks into two types. For the first type, the active set size does not shrink too quickly, i.e.,
$n_j \ge n_{j-1}/(C\alpha)$ for a constant $C$. For each such block~$j$, the bounds
$b_j \le \calO(\alpha) \cdot b_{j-1}$ and $n_j \ge n_{j-1} / \calO(\alpha)$ directly imply that the $j$th term in~\eqref{eq:stochastic-bandits-LHS} is at most $\calO(\alpha^2)$ times the corresponding term in~\eqref{eq:stochastic-bandits-RHS}. For the second type, the active set shrinks rapidly, i.e., $n_j < n_{j-1}/(C\alpha)$. Whenever we have consecutive blocks $j_0, j_0 + 1, \ldots, j_1$ of this type, their contribution to~\eqref{eq:stochastic-bandits-LHS} forms a vanishing geometric series,  and can be charged to the preceding block (i.e., the $(j_0 - 1)$-th term), which is then controlled by $\calO(\alpha^2)$ times the corresponding term in~\eqref{eq:stochastic-bandits-RHS}.

Taking both block types into account, we have
$
    \sum_{j=1}^{B}
    (n_j-n_{j+1})
    \frac{b_j}{n_j}
    \sqrt{\frac{n_{j-1}}{b_{j-1}}}
    \le
    \calO(\alpha^2)
    \sum_{j=1}^{B}
    (n_j-n_{j+1})
    \sqrt{\frac{b_{j-1}}{n_{j-1}}},
$
which establishes the $\calO\left(\alpha^2 \log(NT) \sum_{i:\Delta_i>0} \frac{1}{\Delta_i}\right)$ regret bound. For a general block structure, the algorithm merges consecutive blocks so that the resulting block sequence $\calB'$ satisfies the above geometric-growth condition. The merging only changes the complexity measure $\alpha$ by at most a constant factor, namely,
$\alpha(\mathcal B') \le 4\alpha(\mathcal B)+2$. Therefore, the same regret bound holds for the original adversarially batched instance.
\end{proof}

\noindent\textbf{An Improved Gap-dependent Lower Bound.}
Towards narrowing the gap between the gap-dependent upper and lower bounds, we strengthen the lower bound extended from~\cite{gao2019batched} from $\Omega\left( \alpha \sum_{i:\Delta_i>0} \Delta_i^{-1} \right)$ (\Cref{thm:existing_stochastic_lower}) to $\Omega\left( \alpha^{\frac{3}{2}} \sum_{i:\Delta_i>0} \Delta_i^{-1} \right)$ (\Cref{thm:stochastic_worst_case_lower}) by constructing a carefully designed hard instance. Specifically, we randomly hide a pair of nearly tied Bernoulli arms among \(N\) arms and adopt a three-batch structure of lengths \((N, N^2, N^3)\), so that with constant probability the learner allocates only \(O(N)\) samples to the two special arms before the last batch and hence cannot reliably distinguish the better one, resulting in \(\Omega(N^{5/2})=\Omega\!\left(\alpha^{3/2}\sum_{i:\Delta_i>0}\Delta_i^{-1}\right)\). The full proof of \Cref{thm:stochastic_worst_case_lower} is deferred to \Cref{sec:stochastic_gap_dependent_lower}. 

\paragraph{A Gap-Independent Upper Bound.}
We now prove the gap-independent regret bound of \Cref{algo:stochastic_bandits} (\Cref{thm:stochastic_gap_ind}) via a simple application of the Cauchy-Schwarz inequality \citep{schwarz1890ueber}.
\begin{proof}[Proof of \Cref{thm:stochastic_gap_ind}]
For each arm $i \in [N]$, let $\tau_i$ denote the number of blocks in which it remains active. We will condition on event $\Egood$, under which $\tau_i \le j_i$ holds for every $i \in [N]$.
Then, arm~$i$ is pulled $S_{\tau_i}$ times throughout the algorithm. By definition of the regret, we have
\[
    \Reg_N^{(T)}
    = \sum_{i: \Delta_i > 0}
        \left(\Delta_i\sqrt{S_{\tau_i}}\right)
        \cdot \sqrt{S_{\tau_i}}
    \le \sqrt{\sum_{i: \Delta_i > 0} \Delta_i^2 S_{\tau_i}}
        \cdot \sqrt{\sum_{i: \Delta_i > 0} S_{\tau_i}}
    \le \sqrt{\sum_{i: \Delta_i > 0} \Delta_i^2 S_{\tau_i}}
        \cdot \sqrt{T},
\]
where the second step applies the Cauchy-Schwarz inequality. Let $\pi_1, \pi_2, \ldots, \pi_N \in [N]$ be a permutation of the $N$ arms in the order in which they are eliminated. Formally, we break ties arbitrarily and list all arms that remain active at the end of the list. For each $i \in [N]$, let $\sigma_i \in [N]$ be the index of arm~$i$ in permutation $\pi$. By definition, whenever arm~$i$ remains active in the algorithm, the number of active arms is at least $N - \sigma_i + 1$. For each arm~$i$ with $\Delta_i>0$, we have
\[
    S_{\tau_i}
    = \sum_{k=1}^{\tau_i} \frac{b_k}{n_k}
    \le \frac{\sum_{k=1}^{\tau_i} b_k}{N-\sigma_i+1}
    = \frac{t_{\tau_i}}{N-\sigma_i+1}
    \le \frac{(1+\alpha)t_{\tau_i-1}}{N-\sigma_i+1},
\]
where the last inequality also holds for $\tau_i=1$
under the convention $t_0=1$. Also note that $S_j \geq \sum_{k=1}^j b_k/N = t_j/N$.
If $\tau_i\ge2$, then $\tau_i-1<j_i$, so the definition of $j_i$ gives
\[
    t_{\tau_i-1}
    \le N S_{\tau_i-1}
    < \frac{4\gamma N\ln(NT)}{\Delta_i^2}.
\]
For $\tau_i=1$, the same upper bound on $t_{\tau_i-1}=t_0=1$
holds since $\Delta_i\le1$ and $\gamma$ is sufficiently large.
Therefore, using $\alpha\ge1$, we obtain
$
    S_{\tau_i}
    \le \calO(\alpha)\cdot
    \frac{N\ln(NT)}{(N-\sigma_i+1)\Delta_i^2}.
$
Consequently, on $\Egood$,
\[
    \Reg_N^{(T)}
    \le \sqrt{\sum_{i:\Delta_i>0}\Delta_i^2S_{\tau_i}}
        \cdot\sqrt{T}
    \le \calO(\sqrt{\alpha})\cdot
        \sqrt{T\sum_{i:\Delta_i>0}
        \frac{N\ln(NT)}{N-\sigma_i+1}}.
\]
Finally, $\sum_{i:\Delta_i>0}(N-\sigma_i+1)^{-1}
\le \sum_{i=1}^N 1/i = \calO(\log N)$ gives a bound of $\widetilde{\calO}\left(\sqrt{\alpha TN}\right)$.
Combining with the trivial bound $T$ and $\mathcal{O}(1)$ regret from the complement of $\Egood$ completes the proof.
\end{proof}
\section{An Adaptive Doubling Trick} \label{sec:adaptive_doubling}



To prove \Cref{thm:adaptive_doubling}, we introduce an \emph{adaptive} doubling trick algorithm, as shown in \Cref{algo:doubling}. 
The algorithm maintains an epoch-wise guess $g^{(r)}$ for the sum of squared batch lengths, and runs $\mathcal{A}_{\eta}$ with learning rate $\eta^{(r)}=\sqrt{A/\left(Dg^{(r)}\right)}$. Within each epoch, it tracks the observed quantity $a^{(r)}=\sum b_j^2$ and starts a new epoch once $a^{(r)}\ge g^{(r)}$. The next guess is set to twice the realized sum in epoch~$r$, i.e., $g^{(r+1)} = 2a^{(r)}$. Thereby, the learning rate is recalibrated only when the observed batch complexity outgrows the current guess, yielding regret controlled by the realized squared-batch mass across epochs.

\begin{algorithm}[t]
\caption{Adaptive Learning Rate with Doubling Trick}
\label{algo:doubling}
\begin{algorithmic}
\Require Base algorithm $\mathcal{A}_\eta$ with known regret bound $\mathcal{O} \left(A / \eta + D \eta \|b\|_2^2 \right)$.
\State Play the uniform distribution until the first batch ends (i.e., at time $b_1 = t_1$);
\State $j \gets 1$; $a^{(0)} \gets b_1^2$; \Comment{block~$1$ is regarded as epoch~$0$}
\State $r \gets 1$; $a^{(r)} \gets 0$; \Comment{start epoch~$1$}
\State $g^{(1)} \gets 2a^{(0)}$; $\eta^{(1)} \gets \sqrt{\frac{A}{Dg^{(1)}}}$; \Comment{guess squared $2$-norm in epoch~$1$, set learning rate}
\For{$t = t_1+1, t_1 + 2, \ldots, T$}
    \State Play the action/distribution output by $\mathcal{A}_{\eta}$;
    \If{weights can be updated at time $t$}
        \State $j \gets j + 1$; $t_j \gets t$; $b_j \gets t_j - t_{j-1}$; \Comment{batch~$j$ is completed}
        \State $a^{(r)} \gets a^{(r)} + b_j^2$; \Comment{update squared $2$-norm}
        \State Update $\mathcal{A}_\eta$ with learning rate $\eta^{(r)}$;
        \If{$a^{(r)} \geq g^{(r)}$}
            \State Restart $\mathcal{A}_\eta$;
            \State $r \gets r + 1$; $a^{(r)} \gets 0$; \Comment{start a new epoch~$r$}
            \State $g^{(r)} \gets 2a^{(r - 1)}$; $\eta^{(r)} \gets \sqrt{\dfrac{A}{Dg^{(r)}}}$; \Comment{guess squared $2$-norm, set learning rate}
        \EndIf
    \EndIf
\EndFor
\end{algorithmic}
\end{algorithm}

\begin{proof}[Proof of \Cref{thm:adaptive_doubling}]
Let $R$ be the index of the last non-empty epoch, and let
$\Reg^{(r)}$ denote the regret incurred in epoch
$r \in \{0,1,\ldots,R\}$.
For randomized algorithms, all regret bounds below are understood
in expectation.
By sub-additivity of regret, it suffices to bound each
$\Reg^{(r)}$ separately. Let $L$ denote the upper bound on instantaneous regret implied by the problem setup. Note that $L=\calO(\sqrt{AD})$ by the precondition of the theorem.

\noindent\textbf{Regret in epoch~$0$.}~
Since epoch~$0$ consists of a single block of length $b_1$,
$
\Reg^{(0)}
\le Lb_1
= L\sqrt{a^{(0)}}
= \calO\left(\sqrt{ADa^{(0)}}\right).
$
If $R=0$, this already proves the desired bound.
Henceforth, assume $R\ge1$.

\noindent\textbf{Regret within epochs $1,2,\ldots,R$.}~
Fix $r \in [R]$ and suppose that epoch~$r$ spans blocks
$j_0,j_0+1,\ldots,j_1$.
The regret can be decomposed into two parts:
$\Reg^{(r)} \le \Reg^{(r)}_{\guess}+\Reg^{(r)}_{\remain}$,
where $\Reg^{(r)}_{\guess}$ is the regret in blocks
$j_0$~through~$j_1-1$ and $\Reg^{(r)}_{\remain}$ is the regret
within the last block~$j_1$.
For the first part $\Reg^{(r)}_{\guess}$, note that
\Cref{algo:doubling} guarantees
$\sum_{j=j_0}^{j_1-1}b_j^2<g^{(r)}$; otherwise,
epoch~$r$ should have finished after block~$j_1-1$.
Also recall that \Cref{algo:doubling} sets $\eta^{(r)} = \sqrt{A/(Dg^{(r)})}$.
For a fixed batch structure, the epoch boundaries are deterministic.
Thus, applying the regret guarantee of $\mathcal{A}_{\eta}$
to blocks $j_0$~through~$j_1-1$ gives
\[
\Reg^{(r)}_{\guess}
\le \calO\left(
    \frac{A}{\eta^{(r)}}
    +D\eta^{(r)}\sum_{j=j_0}^{j_1-1}b_j^2
\right)
\le \calO\left(
    \frac{A}{\eta^{(r)}}+D\eta^{(r)}g^{(r)}
\right)
= \calO\left(\sqrt{ADg^{(r)}}\right).
\]
If this prefix is empty, its regret is zero, and the same bound holds.
Since $g^{(r)}=2a^{(r-1)}$, we have
$\Reg^{(r)}_{\guess}
=\calO\left(\sqrt{ADa^{(r-1)}}\right)$. The second part $\Reg^{(r)}_{\remain}$ is upper bounded by
$Lb_{j_1}$.
Since
$a^{(r)}=\sum_{j=j_0}^{j_1}b_j^2\ge b_{j_1}^2$,
we have
$
\Reg^{(r)}_{\remain}
\le L\sqrt{a^{(r)}}
= \calO\left(\sqrt{ADa^{(r)}}\right).
$
Thus,
\[
\Reg^{(r)}
\le \Reg^{(r)}_{\guess}+\Reg^{(r)}_{\remain}
\le \calO\left(
    \sqrt{ADa^{(r-1)}}+\sqrt{ADa^{(r)}}
\right).
\]

\noindent\textbf{Put everything together.}~
We have shown that
$\Reg^{(0)}=\calO\left(\sqrt{ADa^{(0)}}\right)$
and, for each $r\in[R]$,
$
\Reg^{(r)}
\le \calO\left(
    \sqrt{ADa^{(r-1)}}+\sqrt{ADa^{(r)}}
\right).
$
Summing over all epochs shows that the regret of
\Cref{algo:doubling} is upper bounded by
$
\sum_{r=0}^{R}\Reg^{(r)}
\le \sum_{r=0}^{R}\calO\left(\sqrt{ADa^{(r)}}\right)
\le \sum_{r=0}^{R-1}\calO\left(\sqrt{ADa^{(r)}}\right)
    +\calO\left(\sqrt{AD}\|b\|_2\right),
$
where the second step applies
$a^{(R)}\le\sum_{j=1}^{B}b_j^2=\|b\|_2^2$. Recall that \Cref{algo:doubling} sets
$g^{(r)}=2a^{(r-1)}$ for each epoch~$r\in[R]$.
Moreover, for every $r\in[R-1]$,
$a^{(r)}\ge g^{(r)}$ holds when epoch~$r$ ends.
Thus,
$a^{(r)}\ge g^{(r)}=2a^{(r-1)}$,
which further implies
$
a^{(r)}
\le \frac{a^{(R-1)}}{2^{R-r-1}}
$
for every $r\in\{0,1,\ldots,R-1\}$. Therefore,
\begin{align*}
\sum_{r=0}^{R-1}\calO\left(\sqrt{ADa^{(r)}}\right)
\le                 
\calO\left(\sqrt{ADa^{(R-1)}}\right)
\sum_{r=0}^{R-1}\sqrt{\frac{1}{2^{R-r-1}}}
\le \calO\left(\sqrt{ADa^{(R-1)}}\right).
\end{align*}
Noting $\calO\left(\sqrt{ADa^{(R-1)}}\right) \le \calO\left(\sqrt{AD}\|b\|_2\right)$ completes the proof.
\end{proof}

\bibliographystyle{alpha}
\bibliography{reference}

\newcommand{\etalchar}[1]{$^{#1}$}
\begin{thebibliography}{ACBFS02}

\bibitem[ACBF02]{auer2002finite}
Peter Auer, Nicol\`{o} Cesa-Bianchi, and Paul Fischer.
\newblock Finite-time analysis of the multiarmed bandit problem.
\newblock {\em Machine learning}, 47(2):235--256, 2002.

\bibitem[ACBFS02]{auer2002nonstochastic}
Peter Auer, Nicol\`{o} Cesa-Bianchi, Yoav Freund, and Robert~E. Schapire.
\newblock The nonstochastic multiarmed bandit problem.
\newblock {\em SIAM Journal on Computing}, 32(1):48--77, 2002.

\bibitem[BK18]{besson2018doubling}
Lilian Besson and Emilie Kaufmann.
\newblock What doubling tricks can and can't do for multi-armed bandits.
\newblock {\em arXiv preprint arXiv:1803.06971}, 2018.

\bibitem[CBL06]{cesa2006prediction}
Nicol\`{o} Cesa-Bianchi and G{\'a}bor Lugosi.
\newblock {\em Prediction, learning, and games}.
\newblock Cambridge university press, 2006.

\bibitem[CZ23]{chen2023improved}
Mingyu Chen and Xuezhou Zhang.
\newblock Improved algorithms for adversarial bandits with unbounded losses.
\newblock {\em arXiv preprint arXiv:2310.01756}, 2023.

\bibitem[Dru13]{Drucker13}
Andrew Drucker.
\newblock High-confidence predictions under adversarial uncertainty.
\newblock {\em Transactions on Computation Theory}, 5(3):12, 2013.

\bibitem[EKMM21]{esfandiari2021regret}
Hossein Esfandiari, Amin Karbasi, Abbas Mehrabian, and Vahab Mirrokni.
\newblock Regret bounds for batched bandits.
\newblock In {\em AAAI Conference on Artificial Intelligence (AAAI)}, pages
  7340--7348, 2021.

\bibitem[FS97]{freund1997decision}
Yoav Freund and Robert~E Schapire.
\newblock A decision-theoretic generalization of on-line learning and an
  application to boosting.
\newblock {\em Journal of Computer and System Sciences}, 55(1):119--139, 1997.

\bibitem[GHRZ19]{gao2019batched}
Zijun Gao, Yanjun Han, Zhimei Ren, and Zhengqing Zhou.
\newblock Batched multi-armed bandits problem.
\newblock In {\em Advances in Neural Information Processing Systems (NeurIPS)},
  pages 503--513, 2019.

\bibitem[JGS13]{joulani2013online}
Pooria Joulani, Andras Gyorgy, and Csaba Szepesv{\'a}ri.
\newblock Online learning under delayed feedback.
\newblock In {\em International Conference on Machine Learning (ICML)}, pages
  1453--1461, 2013.

\bibitem[JGS16]{joulani2016delay}
Pooria Joulani, Andras Gyorgy, and Csaba Szepesv{\'a}ri.
\newblock Delay-tolerant online convex optimization: Unified analysis and
  adaptive-gradient algorithms.
\newblock In {\em Proceedings of the AAAI Conference on Artificial
  Intelligence}, volume~30, 2016.

\bibitem[JTX{\etalchar{+}}21]{jin2021almost}
Tianyuan Jin, Jing Tang, Pan Xu, Keke Huang, Xiaokui Xiao, and Quanquan Gu.
\newblock Almost optimal anytime algorithm for batched multi-armed bandits.
\newblock In {\em International Conference on Machine Learning}, pages
  5065--5073. PMLR, 2021.

\bibitem[LQ25]{LQ25}
Licheng Liu and Mingda Qiao.
\newblock Online prediction with limited selectivity.
\newblock In {\em Advances in Neural Information Processing Systems (NeurIPS)},
  2025.

\bibitem[LR85]{lai1985asymptotically}
Tze~Leung Lai and Herbert Robbins.
\newblock Asymptotically efficient adaptive allocation rules.
\newblock {\em Advances in applied mathematics}, 6(1):4--22, 1985.

\bibitem[LS20]{LS20}
Tor Lattimore and Csaba Szepesv{\'a}ri.
\newblock {\em Bandit algorithms}.
\newblock Cambridge University Press, 2020.

\bibitem[LW94]{littlestone1994weighted}
Nick Littlestone and Manfred~K Warmuth.
\newblock The weighted majority algorithm.
\newblock {\em Information and computation}, 108(2):212--261, 1994.

\bibitem[Mes05]{mesterharm2005line}
Chris Mesterharm.
\newblock On-line learning with delayed label feedback.
\newblock In {\em International Conference on Algorithmic Learning Theory},
  pages 399--413. Springer, 2005.

\bibitem[OP18]{orabona2018scale}
Francesco Orabona and D{\'a}vid P{\'a}l.
\newblock Scale-free online learning.
\newblock {\em Theoretical Computer Science}, 716:50--69, 2018.

\bibitem[PRCS16]{perchet2016batched}
Vianney Perchet, Philippe Rigollet, Sylvain Chassang, and Erik Snowberg.
\newblock Batched bandit problems.
\newblock {\em The Annals of Statistics}, 44(2):660--681, 2016.

\bibitem[QV19]{QV19}
Mingda Qiao and Gregory Valiant.
\newblock A theory of selective prediction.
\newblock In {\em Conference on Learning Theory (COLT)}, pages 2580--2594,
  2019.

\bibitem[Sch90]{schwarz1890ueber}
Hermann~Amandus Schwarz.
\newblock Ueber ein die fl{\"a}chen kleinsten fl{\"a}cheninhalts betreffendes
  problem der variationsrechnung: Festschrift zum siebzigsten geburtstage des
  herrn karl weierstrass.
\newblock In {\em Gesammelte Mathematische Abhandlungen: Erster Band}, pages
  223--269. Springer, 1890.

\bibitem[SLAM25]{schlisselberg2026improved}
Ofir Schlisselberg, Tal Lancewicki, Peter Auer, and Yishay Mansour.
\newblock Improved best-of-both-worlds regret for bandits with delayed
  feedback.
\newblock {\em Advances in Neural Information Processing Systems},
  38:2035--2072, 2025.

\bibitem[Zin03]{zinkevich2003online}
Martin Zinkevich.
\newblock Online convex programming and generalized infinitesimal gradient
  ascent.
\newblock In {\em Proceedings of the 20th international conference on machine
  learning (icml-03)}, pages 928--936, 2003.

\end{thebibliography}

\clearpage
\appendix
\crefalias{section}{appendix}
\crefalias{subsection}{subappendix}
\crefalias{subsubsection}{subsubappendix}
\startcontents[appendix]
\part*{Appendix}
\printcontents[appendix]{}{1}{}

\clearpage

\section*{AI Use Statement}
We used large language models (LLMs) to assist with writing and editing the manuscript, including improving clarity, grammar, and presentation. The authors reviewed and revised the resulting text and take full responsibility for the content of the paper.

\section{Discussion}
\label{sec:limitation}
A natural open problem is to tighten the gap-dependent regret bounds for stochastic bandits (\Cref{thm:stochastic_bandit,thm:stochastic_worst_case_lower}). While our analysis improved the gap-dependent lower bound from $\Omega\left(\alpha \sum_{i:\Delta_i>0} \Delta_i^{-1}\right)$ to $\Omega\left(\alpha^{\frac{3}{2}}\sum_{i:\Delta_i>0} \Delta_i^{-1}\right)$, there remains an $\alpha^{1/2}$ gap between the upper and lower bounds. 
In \Cref{sec:stochastic_algorithm_lower}, we show a limitation of the comparison used in our current analysis, which suggests that the current approach at best yields an $\widetilde\calO\left( \alpha^{\frac{7}{4}} \sum_{i:\Delta_i>0} \Delta_i^{-1} \right)$ regret bound.
Therefore, one concrete starting point is to generalize such an algorithm-dependent to an algorithm-independent lower bound of $\Omega\left( \alpha^{\frac{7}{4}} \sum_{i:\Delta_i>0} \Delta_i^{-1} \right)$ with a realizable stochastic bandit instance that works for any algorithm, and then see if such a lower bound can be further strengthened to an almost-matching lower bound of $\Omega\left( \alpha^{2} \sum_{i:\Delta_i>0} \Delta_i^{-1} \right)$.

Another interesting direction is to extend adversarially batched online learning beyond the four canonical problems considered in this work. Natural directions include reinforcement learning and other bandit models, such as contextual bandits, where feedback constraints and externally imposed update schedules also arise. The complexity measures developed here, including the $2$-norm of the batch length sequence and the hardness metric $\alpha$, provide a natural starting point for understanding how adversarially chosen batch structures affect regret in these broader online learning models.

\section{Upper Bounds for Stochastic Bandits}
\subsection{Proof of \Cref{lemma:stochastic_est_fail}}
\label{sec:stochastic_est_fail}
\begin{lemma}[Hoeffding's inequality]
\label{lemma:hoeffding}
Let $X_1,\dots,X_\tau$ be independent random variables such that $X_s \in [a,b]$ for all $s \in [\tau]$. Let $\overline{X}=\frac{1}{\tau}\sum_{s=1}^{\tau}X_s$. Then, for any $\epsilon>0$,
\[
\pr{}{\left|\overline{X}-\mathbb{E}[\overline{X}]\right| \geq \epsilon}
\leq
2\exp\left(-\frac{2\tau\epsilon^2}{(b-a)^2}\right).
\]
\end{lemma}

We now prove \Cref{lemma:stochastic_est_fail}. Recall that $j_i = \min\left\{j \in [B]: S_j \ge \frac{4 \gamma \ln(NT)}{\Delta_i^2}\right\}$, and the ``good event'' $\Egood$ is defined as follows: (1) the optimal arm~$i^*$ remains active throughout the algorithm; (2) every sub-optimal arm~$i$ with $\Delta_i>0$ and $j_i \le B$ is eliminated by the end of batch $j_i$.

\begin{proof}[Proof of \Cref{lemma:stochastic_est_fail}]
We first address integer rounding. Under the least-pulled-arm rule, the cumulative pull counts of active arms differ by at most one. Thus, within each block $k$, the number of pulls of any active arm differs from $b_k/n_k$ by at most two. Summing over blocks, the actual cumulative pull count $C_i(j)$ of an arm surviving through block $j$ satisfies $|C_i(j)-S_j|\le 2j$. Since merging produces $O(\log T)$ blocks, choosing $\gamma$ sufficiently large ensures $C_i(j)=\Theta(S_j)$ whenever the elimination threshold is at most one; otherwise, no elimination occurs. We apply concentration inequalities to the actual integer pull counts and suppress these constant-factor rounding effects below.

If Condition~(1) fails, then $\hat{\mu}_i - \hat{\mu}_{i^*} > \sqrt{\frac{\gamma \ln(NT)}{S_j}}$ must hold for some $i \ne i^*$ at the end of some batch $j$.
Since the rewards are bounded in $[0,1]$, $(\hat{\mu}_i-\hat{\mu}_{i^*})-(\mu_i-\mu_{i^*})$ is an average of $S_j$ independent random variables bounded in the length-$4$ interval $[-2, 2]$. Also note that $\mu_i - \mu_{i^*} \le 0$ by the optimality of arm $i^*$. Thus, by \Cref{lemma:hoeffding},
\begin{align*}
    \pr{}{\hat{\mu}_i - \hat{\mu}_{i^*} > \sqrt{\frac{\gamma \ln(NT)}{S_j}}}
&\le\pr{}{(\hat{\mu}_i-\hat{\mu}_{i^*})-(\mu_i-\mu_{i^*}) > \sqrt{\frac{\gamma \ln(NT)}{S_j}}} \\
&\leq 2\exp\left(-\frac{\gamma \ln(NT)}{16}\right) = \frac{2}{(NT)^{\gamma/16}} .
\end{align*}
Union bounding over all possible values of $i \in [N]$, $j \in [B]$, and $S_j \in [T]$ gives
\[
\pr{}{\text{Condition (1) fails}}
\leq
\sum_{i=1}^N \sum_{t=1}^T \sum_{S_j=1}^T
\frac{2}{(NT)^{\gamma/16}}
\leq
\frac{1}{2T},
\]
where the last step holds for some sufficiently large constant $\gamma$.

Next, we analyze the probability that Condition~(2) fails to hold. For this to happen, there must be some arm~$i \ne i^*$ such that $\hat{\mu}_{i^*}-\hat{\mu}_i \leq \sqrt{\frac{\gamma \ln(NT)}{S_{j_i}}}$ by the end of batch~$j_i$. By definition of $j_i = \min\left\{j \in [B]: S_j \geq \frac{4\gamma \ln(NT)}{\Delta_i^2}\right\}$, we have $\mu_{i^*} - \mu_i = \Delta_i \ge 2\sqrt{\frac{\gamma \ln(NT)}{S_{j_i}}}$. It follows that we have
\[
    (\mu_{i^*}-\mu_i)-(\hat{\mu}_{i^*}-\hat{\mu}_i)
\ge \Delta_i-\sqrt{\frac{\gamma \ln(NT)}{S_{j_i}}}
\ge \sqrt{\frac{\gamma \ln(NT)}{S_{j_i}}}.
\]
Applying \Cref{lemma:hoeffding} again upper bounds the probability for the above to hold by $\frac{2}{(NT)^{\gamma/16}}$.
Then, union bounding over all possible values of $i \in [N]$, $j_i \in [B]$, and $S_{j_i} \in [T]$ gives
\[
\pr{}{\text{Condition~(2) fails}}
\leq
NT^2 \cdot
\frac{2}{(NT)^{\gamma/16}}
\leq
\frac{1}{2T}
\]
for some sufficiently large constant $\gamma$. Combining the two probability bounds completes the proof.
\end{proof}

\subsection{Proof of \Cref{lemma:stochastic_cond_regret}}
\label{app:stochastic_cond_regret}
\begin{proof}[Proof of \Cref{lemma:stochastic_cond_regret}]
We first prove the upper bound under the additional assumption that
\begin{equation}\label{eq:assumption}
    b_j \ge 2b_{j-1},~\forall j \in \{2, 3, \ldots, B - 1\}
\quad\text{and}\quad
    b_B \le 2b_{B-1}.
\end{equation}
Later, we will show that \Cref{algo:stochastic_bandits} effectively ensures such an assumption by merging the blocks.

Assuming \Cref{eq:assumption}, it holds for every $j \in [B - 1]$ that
\[
    t_j = b_j + b_{j-1} + \cdots + b_1 \le b_j \cdot \left(1 + \frac{1}{2} + \dots + \frac{1}{2^{j-1}}\right) < 2b_j,
\]
which further implies $\frac{b_j}{b_{j-1}} \le \frac{b_j}{t_{j-1} / 2} \le 2\alpha$ for every $j \in \{2, 3, \ldots, B\}$. Moreover, it holds for every $j \in [B - 1]$ that $S_j = \Theta(b_j / n_j)$, since we have
\[
    \frac{b_j}{n_j} \le S_j = \sum_{k=1}^j \frac{b_k}{n_k}
\le \frac{\sum_{k=1}^j b_k}{n_j}
=   \frac{t_j}{n_j}
\le \frac{2b_j}{n_j}.
\]

\paragraph{Decomposition of the regret.} At the end of each block $j \in [B - 1]$, $n_j - n_{j+1}$ sub-optimal arms are removed. Each such arm~$i \in A_j \setminus A_{j+1}$ was pulled $S_j$ times and each pull caused a $\Delta_{i}$ regret.
We note that, under the least-pulled-arm rule, the cumulative pull counts of active arms differ by at most one. Consequently, the actual cumulative count $C_i(j)$ of an arm surviving through block $j$ satisfies $|C_i(j) - S_j| \leq 2j$. Since merging produces $B = \mathcal{O}(\log T)$ blocks, rounding affects concentration bounds only by constant factors at sample sizes where elimination can occur. Its contribution to regret therefore is $\mathcal{O}(\log T \sum_{i:\Delta_i>0} \Delta_i)$, which is absorbed by the stated bounds. 

Conditioned on event $\Egood$, every sub-optimal arm~$i$ is removed before or at the end of block~$j_i$. Then, $i \in A_j$ implies that
\[
    \Delta_i \leq \sqrt{\frac{\gamma \ln(NT)}{S_{j-1}}}.
\]
The above also holds for the last block (i.e., $j = B$), if we define $A_{B+1} \coloneqq \{i \in [N]: \Delta_i = 0\}$ as the set of optimal arms and $n_{B+1} = |A_{B+1}|$.

Summing over all $B$ blocks, we obtain the following upper bound on the regret:
\[
    \sum_{j=1}^{B} (n_j - n_{j+1}) \cdot S_j \cdot \sqrt{\frac{\gamma \ln(NT)}{S_{j-1}}}
=   \calO(\sqrt{\log(NT)}) \cdot \sum_{j=1}^{B} (n_j - n_{j+1}) \cdot S_j \cdot \sqrt{\frac{1}{S_{j-1}}}.
\]
Recall that $S_j = \Theta(b_j / n_j)$ for every $j \in [B - 1]$. Therefore, using the shorthand
\[
    \LHS_j
\coloneqq \begin{cases}
    (n_j - n_{j+1}) \cdot \frac{b_j}{n_j} \cdot \sqrt{\frac{n_{j-1}}{b_{j-1}}}, & j \in [B - 1],\\
    (n_j - n_{j+1}) \cdot S_j \cdot \sqrt{\frac{n_{j-1}}{b_{j-1}}}, & j = B,
    \end{cases}
\]
the regret is upper bounded by $\calO(\sqrt{\log(NT)}) \cdot \sum_{j=1}^{B}\LHS_j$.

\paragraph{Decomposition of the regret bound.} Recall that our goal is to upper bound the regret by 
\begin{align*}
    \alpha^2\ln(NT) \cdot \sum_{i: \Delta_i > 0} \frac{1}{\Delta_i}
&\ge \alpha^2 \ln(NT) \sum_{j=1}^{B} (n_j - n_{j+1}) \sqrt{\frac{S_{j-1}}{\gamma\log(NT)}}\\
&\ge \Omega(\sqrt{\log(NT)}) \cdot \alpha^2\sum_{j=1}^{B} (n_j - n_{j+1}) \sqrt{\frac{b_{j-1}}{n_{j-1}}}.
\end{align*}
Therefore, it suffices to show that $\sum_{j=1}^{B} \LHS_j \leq \calO(\alpha^2) \cdot \sum_{j=1}^{B}\RHS_j$,
where
\[
    \RHS_j \coloneqq (n_j - n_{j+1}) \sqrt{\frac{b_{j-1}}{n_{j-1}}}.
\]


\paragraph{Division of blocks.} Towards proving $\sum_{j=1}^{B} \LHS_j \leq \calO(\alpha^2) \cdot \sum_{j=1}^{B} \RHS_j$, we set $C = 3$ and partition the blocks $1, 2, \ldots, B$ into two sets:
\[
    J_1 \coloneqq \left\{j \in [B]: n_j \geq \frac{n_{j-1}}{C\alpha}\right\}
\quad\text{and}\quad
    J_2 \coloneqq \left\{j \in [B]: n_j < \frac{n_{j-1}}{C\alpha}\right\}.
\]

For notational convenience, we extend the initialization convention $t_0=1$ by setting $b_0:=1$, $n_0:=n_1=N$, and $S_0:=b_0/n_0$. These analysis-only boundary values encode the trivial first-block bounds $\Delta_i\le1\le\sqrt{\gamma\log(NT)/S_0}$ and $b_1\le\alpha b_0$, while ensuring the first block belongs to the $J_1$.

\paragraph{Blocks in $J_1$.} For any $j \in J_1$, the condition $n_j \geq \frac{n_{j-1}}{C\alpha}$ together with the bound $b_j \le 2\alpha \cdot b_{j-1}$ gives
\[
    \LHS_j
\le (n_j - n_{j+1}) \cdot \frac{2\alpha \cdot b_{j-1}}{n_{j-1} / (C\alpha)} \cdot \sqrt{\frac{n_{j-1}}{b_{j-1}}}
\le \calO(\alpha^2) \cdot \RHS_j,
\]
which further implies
\begin{equation}\label{eq:J1-bound}
    \sum_{j \in J_1}\LHS_j
\le \calO(\alpha^2) \cdot \sum_{j \in J_1}\RHS_j
\le \calO(\alpha^2) \cdot \sum_{j=1}^{B}\RHS_j.
\end{equation}

\paragraph{Blocks in $J_2$.} To control the contribution from $J_2$, consider an index pair $(j_0, j_1)$ that specify a maximal contiguous subsequence of indices in $J_2$. In other words, $j_0, j_0 + 1, \ldots, j_1 \in J_2$ while $j_0 - 1, j_1 + 1 \notin J_2$. We will show that $\sum_{j=j_0}^{j_1} \LHS_j \leq \calO(\alpha^2) \cdot \RHS_{j_0 - 1}$.

When $j \ne B$, we have
\[
    \LHS_j
=   (n_j - n_{j+1}) \cdot \frac{b_j}{n_j} \sqrt{\frac{n_{j-1}}{b_{j-1}}}
\leq b_j \sqrt{\frac{n_{j-1}}{b_{j-1}}}
\leq \calO(\alpha) \sqrt{n_{j-1} b_{j-1}},
\]
where the last step applies $b_j \le 2\alpha \cdot b_{j-1}$. For $j = B$, note that $S_{B-1} \le 2b_{B-1} / n_{B-1}$ and the assumption $b_B \le 2b_{B-1}$ (from \Cref{eq:assumption}) together imply
\[
    S_B
=   S_{B-1} + \frac{b_B}{n_B}
\le \frac{2b_{B-1}}{n_{B-1}} + \frac{2b_{B-1}}{n_B}
\le \frac{4b_{B-1}}{n_B}.
\]
It follows that $\LHS_B = (n_B - n_{B+1}) \cdot S_B \sqrt{\frac{n_{B-1}}{b_{B-1}}} \le 4\sqrt{n_{B-1}b_{B-1}}$. Therefore, we have $\LHS_j \le \calO(\alpha) \sqrt{n_{j-1} b_{j-1}}$ for every $j \in \{j_0, j_0 + 1, \ldots, j_1\}$.

Since $b_j \leq 2\alpha \cdot b_{j-1}$ and $n_j < \frac{n_{j-1}}{C\alpha}$ hold for every $j \in \{j_0, j_0 + 1, \ldots, j_1\}$, a simple induction gives $b_{j-1} \le (2\alpha)^{j - j_0} \cdot b_{j_0 - 1}$ and $n_{j-1} \le \frac{n_{j_0 - 1}}{(C\alpha)^{j - j_0}}$. It follows that
\begin{align*}
    \sum_{j=j_0}^{j_1} \LHS_j
&\le \calO(\alpha) \cdot \sum_{j=j_0}^{j_1} \sqrt{b_{j-1}n_{j-1}} \tag{$\LHS_j \le \calO(\alpha) \sqrt{n_{j-1} b_{j-1}}$}\\
&\le \calO(\alpha) \cdot \sum_{j=j_0}^{j_1} \sqrt{(2\alpha)^{j - j_0} \cdot b_{j_0 - 1} \cdot \frac{n_{j_0 - 1}}{(C\alpha)^{j - j_0}}}\\
&\le \calO(\alpha) \cdot \sqrt{b_{j_0 - 1} n_{j_0 - 1}} \cdot \sum_{t=0}^{j_1 - j_0}\sqrt{(2/3)^t}
=  \calO(\alpha) \cdot \sqrt{b_{j_0 - 1} n_{j_0 - 1}}. \tag{$C = 3$}
\end{align*}
For the $\RHS_{j_0 - 1}$ term, 
note that the assumption $j_0 \in J_2$ implies $n_{j_0} < \frac{n_{j_0-1}}{C\alpha}$. Moreover, since $j_0 - 1 \in J_1$, we have $n_{j_0 - 1} \geq \frac{n_{j_0 - 2}}{C \alpha}$ and equivalently $n_{j_0 - 2} \leq C\alpha \cdot n_{j_0 - 1}$. We then have 
\begin{align*}
    \RHS_{j_0 - 1}
&=  (n_{j_0-1} - n_{j_0})\sqrt{\frac{b_{j_0 - 2}}{n_{j_0 - 2}}}\\
&>  \left(n_{j_0 - 1} - \frac{n_{j_0 - 1}}{C\alpha}\right) \sqrt{\frac{b_{j_0 - 2}}{C\alpha \cdot n_{j_0 - 1}}} \tag{$n_{j_0} < \frac{n_{j_0-1}}{C\alpha}$ and $n_{j_0 - 2} \le C\alpha \cdot n_{j_0 - 1}$}\\
&\ge\Omega(1) \cdot \sqrt{\frac{b_{j_0 - 2} n_{j_0 -1}}{\alpha}}.
\end{align*}

Combining the two bounds above and applying $b_{j_0 - 1} \le 2\alpha \cdot b_{j_0 - 2}$, we conclude that
\[
    \sum_{j=j_0}^{j_1} \LHS_j
\le \calO(\alpha) \cdot \sqrt{b_{j_0 - 1} n_{j_0 - 1}}
\le \calO(\alpha) \cdot \sqrt{\alpha \cdot b_{j_0 - 2} n_{j_0 - 1}}
\le \calO(\alpha^2) \cdot \RHS_{j_0 - 1}.
\]

Summing over all contiguous intervals in $J_2$ (i.e., all pairs $(j_0, j_1)$) gives
\begin{equation}\label{eq:J2-bound}
    \sum_{j \in J_2}\LHS_j \le \calO(\alpha^2) \cdot \sum_{j=1}^{B}\RHS_j.
\end{equation}

Combining \Cref{eq:J1-bound,eq:J2-bound} completes the proof under the assumption in \Cref{eq:assumption}. 

\paragraph{General block structures.} Finally, we note that \Cref{algo:stochastic_bandits} effectively merges the blocks to form a new batch structure, denoted as $\calB'$, that ensures $b'_j \geq 2b'_{j-1}$ except for the last block. Since the assumption in \Cref{eq:assumption} holds, our previous analysis bounds the regret by $\calO\left([\alpha(\calB')]^2 \log(NT) \sum_{i: \Delta_i > 0} \frac{1}{\Delta_i}\right)$. It remains to show that the merging does not significantly increase the difficulty parameter $\alpha$ of the batch structure. Concretely, we will prove $\alpha(\calB') \le 4\alpha(\calB) + 2$.

Consider an arbitrary block $b'_j$ in the new batch structure $\calB'$. 
If $b'_j$ is the final incomplete merged block, we have
$b'_j < 2b'_{j-1} \le 2t'_{j-1}$.
Otherwise, $b'_j$ is a completed merged block, including the case
where its completion coincides with the end of the horizon.
In this boundary case, the final-block condition in Equation~(3)
is replaced by $b_B \ge 2b_{B-1}$; all merged blocks then satisfy
the geometric-growth condition, so $S_B=\Theta(b_B/n_B)$ and
the regular-block estimates in the preceding regret analysis
apply to the final block as well.
For a completed merged block $b'_j$, let $r$ be the index of
the last original block (in $\calB$) that forms $b'_j$.
By construction, we have $b'_j \le 2b'_{j-1} + b_r$; otherwise, block $b'_j$ should be completed after merging block $b_{r-1}$. By definition, we have $b_r / t_{r-1} \le \alpha(\calB)$. Furthermore,
\[
    t_{r-1}
\le t'_{j-1} + 2b'_{j-1}
=   b'_1 + b'_2 + \cdots + b'_{j-1} + 2b'_{j-1}
\le 4b'_{j-1}.
\]
Then, we have
\[
    b'_j
\le 2b'_{j-1} + \alpha(\calB) \cdot t_{r-1}
\le 2b'_{j-1} + \alpha(\calB) \cdot 4b'_{j-1}
\le (4\alpha(\calB) + 2) \cdot t'_{j-1}.
\]
Therefore, we have $\alpha(\calB') = \max_j \frac{b'_j}{t'_{j-1}} \le 4\alpha(\calB) + 2$, which completes the proof.
\end{proof}

\subsection{Extension of Existing Bounds for Arbitrary Batch Structure}
\label{sec:stochastic_alpha_bound}
\begin{proposition}
\label{propo:existing_alpha_bound}
For the adversarially batched stochastic bandits with $N$ arms and arbitrary batch structure $\calB$ characterized by the hardness metric $\alpha$, there is an algorithm with an expected regret of
\[
\calO\left( \alpha \log N \log(NT) \frac{N}{\Delta_{\min}} \right),
\]
where $\Delta_{\min} = \min_{i: \Delta_i > 0} \Delta_i$.
\end{proposition}
\begin{proof}
We follow the argument used in the proof of \Cref{thm:stochastic_gap_ind}.
Throughout the following argument, we work on the event $\Egood$.
Let $\tau_i$ and $\sigma_i$ be defined as in that proof:
$\tau_i$ is the number of blocks in which arm $i$ remains active,
and $\sigma_i$ is its position in the elimination order, with ties
broken arbitrarily and surviving arms listed last.
Then $\tau_i \leq j_i$, and
\[
\Reg_N^{(t_B)}
= \sum_{i: \Delta_i > 0} \Delta_i \cdot S_{\tau_i}.
\]
Whenever arm $i$ remains active, there are at least
$N - \sigma_i + 1$ active arms. Hence,
\[
S_{\tau_i}
= \sum_{k=1}^{\tau_i} \frac{b_k}{n_k}
\leq \sum_{k=1}^{\tau_i}\frac{b_k}{N - \sigma_i + 1}
= \frac{t_{\tau_i}}{N - \sigma_i + 1}
\leq \frac{(1+\alpha)t_{\tau_i - 1}}{N - \sigma_i + 1}.
\]
Since $\tau_i \leq j_i$, the definition of $j_i$ gives, for
$\tau_i \geq 2$,
\[
\frac{t_{\tau_i - 1}}{N}
\leq S_{\tau_i - 1}
< \frac{4\gamma\log(NT)}{\Delta_i^2}.
\]
Consequently,
\[
t_{\tau_i - 1}
\leq \frac{4\gamma\log(NT)N}{\Delta_i^2}.
\]
The same bound holds for $\tau_i=1$, using $t_0=1$ and the
choice of $\gamma$ in \Cref{lemma:stochastic_est_fail}.
Moreover,
\[
\sum_{i: \Delta_i > 0} \frac{1}{N - \sigma_i + 1}
\leq \sum_{n=1}^N \frac{1}{n}
\leq 1+\ln N,
\]
since $(\sigma_i)_{i\in[N]}$ is a permutation of $[N]$.
Combining these bounds, we have
\begin{align*}
\Reg_N^{(t_B)}
&= \sum_{i: \Delta_i > 0} \Delta_i \cdot S_{\tau_i}\\
&\leq \sum_{i: \Delta_i > 0}
\Delta_i \frac{(1+\alpha)t_{\tau_i - 1}}{N - \sigma_i + 1}\\
&\leq \sum_{i: \Delta_i > 0}
\Delta_i \frac{4\gamma\log(NT)N}{\Delta_i^2}
\frac{1+\alpha}{N - \sigma_i + 1}\\
&\leq 4\gamma(1+\alpha)(1+\ln N)\log(NT)
\frac{N}{\Delta_{\min}}.
\end{align*}
By \Cref{lemma:stochastic_est_fail}, $\Pr[\Egood]\geq 1 - 1/T$.
Since the regret is at most $T=t_B$, the contribution of
bad events to the expected regret is at most $1$.
Using $\alpha\geq 1$ and $N\geq 2$, we obtain
\[
\mathbb{E}\!\left[\Reg_N^{(t_B)}\right]
\leq 4\gamma(1+\alpha)(1+\ln N)\log(NT)
\frac{N}{\Delta_{\min}}+1
= \mathcal{O}\!\left(
\alpha\log N\log(NT)\frac{N}{\Delta_{\min}}
\right),
\]
which completes the proof.
\end{proof}

\section{Lower Bounds for Stochastic Bandits}
\subsection{Lower Bounds Extended from \cite{gao2019batched}}
\label{app:existing_stochastic_lower}
\paragraph{Gap-Dependent Lower Bound.}
We first give a bounded-reward analogue of \cite[Lemma 2]{gao2019batched}. The proof follows the same multiple-hypothesis testing argument, but uses Bernoulli reward distributions instead of Gaussian ones.
\begin{lemma}
\label{lemma:stochastic_bernoulli_lower_bound}
For every $N \geq 2$ and every $0 < \Delta \leq 1/8$, and every batched policy, there exists an $N$-armed stochastic bandit instance with Bernoulli rewards, with all means in $[0, 1]$ and all nonzero gaps belonging to $\{ \Delta, 2\Delta \}$, such that every batched policy satisfies
\[
\mathbb{E}[\mathrm{Reg}] \geq \frac{\Delta}{4} \sum_{j=1}^B b_j \exp \left( -C_0 \cdot \frac{t_{j-1}\Delta^2}{N - 1} \right),
\]
where $C_0 > 0$ is a universal numerical constant. 
\end{lemma}
\begin{proof}
Fix $\Delta \leq 1/8$. We construct $N$ candidate bandit instances $P_1, \dots, P_N$. Under $P_1$, the arm reward distributions are 
\[
P_1 = \mathrm{Bern}\left(\frac{1}{2}+\Delta\right)
\otimes
\mathrm{Bern}\left(\frac{1}{2}\right)^{\otimes (N-1)}.
\]
For every $i \in \{ 2, \dots, N \}$, define
\[
P_i = \mathrm{Bern} \left( \frac{1}{2} + \Delta \right) \otimes \mathrm{Bern}\left(\frac{1}{2}\right) \otimes \dots \otimes \mathrm{Bern}\left(\frac{1}{2} + 2\Delta\right) \otimes \dots \otimes \mathrm{Bern}\left(\frac{1}{2}\right),
\]
where the $i$-th arm has a mean of $1/2 + 2\Delta$. As $\Delta \leq 1/8$, 
\[
1/2 \leq 1/2+\Delta \leq 1/2 + 2\Delta \leq 3/4,
\]
so every reward distribution is supported on $\{ 0, 1 \}$ and has mean in $[0, 1]$. Moreover, under $P_1$, arm~$1$ is optimal, and every other arm has gap exactly $\Delta$. Under $P_i$ for each $i\ge2$, arm~$i$ is optimal; arm~$1$ has gap $\Delta$, while every arm~$k\notin\{1,i\}$ has gap $2\Delta$. Hence every nonzero gap belongs to $\{\Delta,2\Delta\}$, and in particular is at most $2\Delta \leq 1/4$. 

For a round $t$ belonging to batch $j$, let $P_i^t$ denote the distribution, under instance $P_i$, of all information available to the learner when the action at time $t$ is selected. Since no feedback from the current batch is available before the batch ends, $P_i^t$ depends only on observations from the first $j-1$ batches. Every incorrect arm choice incurs instantaneous regret of at least $\Delta$ under each $P_i$. Therefore, 
\[
\sup_{i \in [N]} \mathbb{E}_i [\mathrm{Reg}] \geq \frac{1}{N} \sum_{i=1}^N \mathbb{E}_i [\mathrm{Reg}] \geq \Delta \sum_{t=1}^T \frac{1}{N} \sum_{i=1}^N P_i^t (i_t \neq i).
\]
We apply the same multiple-hypothesis testing inequality used in \cite{gao2019batched} with the star tree having edges
$
(1, 2), (1, 3), \dots, (1, N).
$
For every $t$, 
\[
\frac{1}{N} \sum_{i=1}^N P_i^t (i_t \neq i) \geq \frac{1}{2N} \sum_{i=2}^N \exp(-D_{\mathrm{KL}}(P_1^t \| P_i^t)).
\]
The instances $P_1$ and $P_i$ differ only in the reward distribution of arm~$i$: under $P_1$ it is $\operatorname{Bern}(1/2)$, whereas under $P_i$ it is $\operatorname{Bern}(1/2+2\Delta)$. Hence, the standard KL chain rule for bandit observations gives
\[
D_{\mathrm{KL}}(P_1^t\Vert P_i^t)=\mathbb E_1[t_i]\,d_\Delta,
\tag{22}
\]
where $t_i$ is the number of times arm~$i$ has been pulled before the current batch and
\begin{align*}
d_\Delta &= D_{\mathrm{KL}}\left(\operatorname{Bern}\left(1/2\right)\middle\Vert \operatorname{Bern}\left(1/2+2\Delta\right)\right)\\
&=\frac12\log\frac{1/2}{1/2+2\Delta}+\frac12\log\frac{1/2}{1/2-2\Delta}\\
&=-\frac12\log(1-16\Delta^2).
\end{align*}
Since $\Delta \leq 1/8$, we have $16 \Delta^2 \leq 1/4$. Using 
\[
-\log(1-x) \leq \frac{x}{1-x},
\]
where $0 \leq x < 1$, we obtain 
\[
d_\Delta \leq \frac{1}{2} \frac{16\Delta^2}{1-16\Delta^2} \leq C_0 \Delta^2.
\]
For $t$ in batch $j$, all observations available to the learner come from the first $t_{j-1}$ rounds. Consequently, 
\[
\sum_{i=2}^N \mathbb{E}_1 [t_i] \leq t_{j-1}.
\]
Together, by Jensen's inequality, we have 
\begin{align*}
    \frac{1}{N} \sum_{i=1}^N P_i^t (i_t \neq i) &\geq \frac{1}{2N} \sum_{i=2}^N \exp(-d_\Delta \mathbb{E}_1[t_i])\\
    &\geq \frac{N-1}{2N} \exp \left( -\frac{d_\Delta}{N-1} \sum_{i=2}^N \mathbb{E}_1 [t_i] \right)\\
    &\geq \frac{1}{4} \exp \left( -C_0 \frac{t_{j-1}\Delta^2}{N-1} \right)
\end{align*}
Summing over all $b_j$ rounds in batch $j$ gives 
\[
\sup_{i \in [N]} \mathbb{E}_i [\mathrm{Reg}] \geq \frac{\Delta}{4} \sum_{j=1}^B b_j \exp \left( -C_0 \frac{t_{j-1}\Delta^2}{N-1} \right),
\]
which marks the end of the proof.
\end{proof}
Now we are ready to prove \Cref{thm:existing_stochastic_lower}. 
\begin{proof}[Proof of \Cref{thm:existing_stochastic_lower}]
We consider a broad family of hard instances that satisfy $t_{j^* - 1} \geq CN$. Taking advantage of \Cref{lemma:stochastic_bernoulli_lower_bound}, we have 
\begin{align*}
    \Delta \cdot \Ex{}{\Reg_N^{(B)}}
&\geq \Delta^2 \cdot \sum_{j=1}^{B} \frac{t_j - t_{j-1}}{4} \exp\!\left( -\frac{C_0 t_{j-1}\Delta^2}{N-1} \right)\\
&= \Delta^2 \cdot \sum_{j=1}^{B} \frac{b_j}{4} \exp\!\left( -\frac{C_0 t_{j-1}\Delta^2}{N-1} \right)\\
&\geq \Delta^2 \cdot \frac{b_{j^*}}{4} \exp\!\left( -\frac{C_0 t_{j^*-1}\Delta^2}{N-1} \right),
\end{align*}
where we set $j^*$ to be the block index chosen by $\alpha$, i.e., $\alpha = \max_j \frac{b_j}{t_{j-1}} = \frac{b_{j^*}}{t_{j^* - 1}}$. Next, we consider a case where every sub-optimal arm has the smallest gap $\Delta = \Delta_{i: i \neq i^*} = \sqrt{\frac{N-1}{C_0 t_{j^*-1}}}$, which leads to
\begin{align*}
\Delta \cdot  \Ex{}{\Reg_N^{(B)}} &\geq \Delta^2 \cdot \frac{b_{j^*}}{4} \exp\!\left( -\frac{C_0 t_{j^*-1}\Delta^2}{N-1} \right)\\
&\geq \frac{N - 1}{t_{j^* - 1}} \frac{b_{j^*}}{4} \exp \left( -\frac{C_0 t_{j^* - 1}}{N - 1} \frac{N - 1}{t_{j^* - 1}} \right)\\
&= \Omega(1) \cdot \frac{b_{j^*}}{t_{j^* - 1}} N = \Omega(1) \cdot \alpha N.
\end{align*}
We note that, based on the hard instance construction ($t_{j^* - 1} \geq CN$), $\Delta =  \sqrt{\frac{N-1}{C_0t_{j^* - 1}}} \leq 1/8$ is a valid bounded-reward gap. Finally, we have 
\[
\Delta \cdot \Ex{}{\Reg_N^{(B)}} \geq \Omega(1) \cdot N \alpha \iff \Ex{}{\Reg_N^{(B)}} \geq \Omega(1) \cdot \alpha\frac{N}{\Delta} = \Omega \left( \alpha \sum_{i \neq i^*} \Delta_i^{-1} \right),
\]
which completes the proof.
\end{proof}

\paragraph{Gap-Independent Lower Bound.}
Similar to the proof of \Cref{thm:existing_stochastic_lower}, we consider a family of hard instances that satisfy $t_{j^* - 1} \geq CN$ and $t_{j^* - 1} \geq \Omega(T/\alpha)$, where, intuitively speaking, the second part asks that the hard instance's largest relative batch must occur late enough. Based on the proof of \Cref{thm:existing_stochastic_lower}, we have
\begin{align*}
    \Ex{}{\Reg_N^{(B)}} &\geq \Delta \cdot \sum_{j=1}^{B} \frac{t_j - t_{j-1}}{4} \exp\!\left( -\frac{C_0 t_{j-1}\Delta^2}{N-1} \right)\\
    &= \Delta \cdot \sum_{j=1}^{B} \frac{b_j}{4} \exp\!\left( -\frac{C_0 t_{j-1}\Delta^2}{N-1} \right)\\
    &\geq \frac{\Delta}{4} b_{j^*} \exp \left( -\frac{C_0 t_{j^*-1}\Delta^2}{N-1} \right)\\
    &= \frac{e^{-1}}{4\sqrt{C_0}} b_{j^*} \sqrt{\frac{N-1}{t_{j^*-1}}}.
\end{align*}
Recall $b_{j^*} = \Theta (\alpha t_{j^* - 1})$ and the condition of $t_{j^* - 1} = \Omega\left( \frac{T}{\alpha} \right)$, we have 
\[
\Ex{}{\Reg_N^{(B)}} \geq \Omega(1) \cdot \alpha \sqrt{N t_{j^* - 1}} \geq \Omega(1) \cdot \sqrt{\alpha NT}.
\]

\subsection{An Improved Gap-Dependent Lower Bound}
\label{sec:stochastic_gap_dependent_lower}
\begin{proof}
Set $\delta = \frac{1}{16\sqrt{N}}$. For every unordered pair $\{ u, v \} \subseteq [N]$, define two instances $\nu_{u, v}^+$ and $\nu_{u, v}^-$. Under $\nu_{u, v}^+$, 
\[
\mu_u = \frac{1}{2} + \frac{\delta}{2},\; \mu_v = \frac{1}{2} - \frac{\delta}{2},
\]
while every arm~$i \notin \{ u, v \}$ has deterministic reward $0$. Under $\nu_{u, v}^-$, the means of $u$ and $v$ are exchanged, i.e., 
\[
\mu_u = \frac{1}{2} - \frac{\delta}{2},\; \mu_v = \frac{1}{2} + \frac{\delta}{2}.
\]
The two special arms have Bernoulli rewards.

Our hard distribution samples $\{U,V\} \sim \operatorname{Unif}\left(\binom{[N]}{2}\right)$, and independently samples $S \in \{ +, -\}$ uniformly. The resulting instance is $\nu_{U, V}^S$. The batch structure is set to be $\mathcal{B} = \{ N, N^2, N^3\}$, which gives $\alpha = \max \left\{ N, \frac{N^2}{N}, \frac{N^3}{N + N^2} \right\} = \Theta(N)$. We first focus on the deterministic algorithm and extend to the randomized algorithm at the end. 

\paragraph{The First Batch.}
Recall that the first batch has $N$ rounds without any prior feedback. Let $m_i$ be the number of times that arm~$i$ has been pulled in the first batch; then $\sum_{i=1}^N m_i = N$. Define the set of arms that are not heavily sampled (less than 3 times) in the first batch as $\mathcal{L} = \{ i \in [N]: m_i \leq 2 \}$. Since every arm outside $\mathcal{L}$ is pulled at least three times, we have 
\[
3(N - |\mathcal{L}|) \leq \sum_{i=1}^N m_i = N,
\]
and hence $|\mathcal{L}| \geq \frac{2N}{3}$. Let $E$ be the event that all rewards received in the first batch are $0$, and define $F = E \cap \{U, V \in \mathcal{L}\}$. Recall that $\mu_i \leq \frac{1}{2} + \frac{\delta}{2} < \frac{1}{2} + \frac{1}{4} = \frac{3}{4}$ as $\delta = \frac{1}{16\sqrt{N}} < 1/2$. For any $u, v \in \mathcal{L}$ and either $S=+$ or $S=-$, both special arms have means smaller than $3/4$. Therefore, 
\[
\pr{}{r_i=0 \mid i \in \{u, v\}} \geq \frac{1}{4}.
\]
Moreover, $m_u + m_v \leq 4$. Thus,
\[
\pr{}{ E \mid U = u, V=v, S } \geq 4^{-(m_u + m_v)} \geq 4^{-4} = \Omega(1).
\]
Consequently,
\[
\Pr[F] = \Pr[U, V \in \mathcal{L}] \cdot \Pr[E \mid  U, V \in \mathcal{L}] \geq \frac{\binom{|\mathcal{L}|}{2}}{\binom{N}{2}}4^{-4} = \Omega(1).
\]

\paragraph{The Second Batch.}
Conditioned on event $E$, the learner observes the same all-zero history regardless of the underlying pair and sign. Let $q_i$ be the number of pulls of arm~$i$ in the second batch. Since the algorithm is deterministic, $\sum_{i=1}^N q_i = N^2$, and the values $q_1, \dots, q_N$ are fixed conditioned on $E$. For $u, v \in \mathcal{L}$, we denote $p_{u,v} = \Pr[E \mid U=u, V=v]$, where the probability averages over the random sign $S$. By the argument above, $p_{u,v} \geq 4^{-4}$. We check the expected sample complexity of special arms in the second batch conditioning on $F$:
\[
\mathbb{E}[q_U + q_V | F] = \frac{\sum_{u < v;\,u, v \in \mathcal{L}} p_{u,v} \cdot (q_u + q_v)}{\sum_{u < v;\;u,v \in \mathcal{L}} p_{u,v}} \leq \frac{\sum_{u < v;\;u,v \in \mathcal{L}} (q_u + q_v)}{\sum_{u < v;\,u,v \in \mathcal{L}} p_{u,v}}
\]
as $p_{u,v} \leq 1$. Moreover, recall that $p_{u,v} \geq 4^{-4}$ and note that every $q_i$ will appear in $(q_u + q_v)$ with a total number of $|\mathcal{L}| - 1$ times. Thus,
\[
\frac{\sum_{u < v;\,u,v\in \mathcal{L}} (q_u + q_v)}{\sum_{u < v;\,u,v \in \mathcal{L}} p_{u,v}} \leq \frac{(|\mathcal{L}|-1) \sum_{i\in \mathcal{L}} q_i}{4^{-4} \binom{|\mathcal{L}|}{2}} \leq \frac{(|\mathcal{L}|-1) N^2}{4^{-4} \binom{|\mathcal{L}|}{2}} = \mathcal{O}(N),
\]
where the last step follows from $|\mathcal{L}| = \Omega(N)$.
\paragraph{The Third Batch.}
Next, conditioned on event $F$, we fix one $(u, v)$ pair with $u,v \in \mathcal{L}$ form a binary hypothesis testing:
\[
H_+: \mu_u = \frac{1}{2} + \frac{\delta}{2}; \; \mu_v = \frac{1}{2} - \frac{\delta}{2},
\]
and 
\[
H_-: \mu_u = \frac{1}{2} - \frac{\delta}{2}; \; \mu_v = \frac{1}{2} + \frac{\delta}{2}.
\]
We first note that conditioning on the first-batch event $E$ does not make the two hypotheses equally likely in general. Nevertheless, since $u,v \in \mathcal{L}$, their posterior probabilities remain bounded away from zero. Indeed, 
\[
\frac{\pr{}{E \mid U = u, V=v, S=+}}{\pr{}{ E \mid U = u, V=v, S=- }} = \frac{\left( \frac{1-\delta}{2} \right)^{m_u} \left( \frac{1+\delta}{2} \right)^{m_v}}{\left( \frac{1+\delta}{2} \right)^{m_u} \left( \frac{1-\delta}{2} \right)^{m_v}} = \left( \frac{1-\delta}{1+\delta} \right)^{m_u - m_v}.
\]
Since $m_u, m_v \leq 2$ and $\delta \leq 1/16$, this likelihood ratio is bounded above and below by universal constants. In other words, there exists a universal constant $c_0 > 0$ such that 
\[
\pr{}{S=+|E,U=u, V=v} \geq c_0,\;\; \pr{}{S=-|E,U=u, V=v} \geq c_0.
\]
Let $P_+$ and $P_-$ denote the distribution of the second-batch observations under $H_+$ and $H_-$ conditioned on the first-batch history $E$. Since the second-batch allocation is fixed to be $q_1, \dots, q_N$, the KL divergence between the distributions induced by the two hypotheses can be formulated as
\[
D_{\mathrm{KL}} (P_{+} \| P_{-}) = q_u \cdot \mathrm{kl}\left(\frac{1+\delta}{2}, \frac{1-\delta}{2}\right) + q_v \cdot \mathrm{kl}\left(\frac{1-\delta}{2}, \frac{1+\delta}{2}\right),
\]
where $\mathrm{kl}(p\|q)$ denotes the binary relative entropy. It is easy to see that 
\[
\mathrm{kl}\left(\frac{1+\delta}{2}, \frac{1-\delta}{2}\right) = \mathrm{kl}\left(\frac{1-\delta}{2}, \frac{1+\delta}{2}\right) = \delta \log \frac{1 + \delta}{1 - \delta} \leq \delta \cdot 4\delta  = 4 \delta^2.
\]
Thus, $\mathrm{KL} (P_{+} \| P_{-}) \leq 4\delta^2 (q_u + q_v)$. Let $X_i$ denote the number of times that arm $i$ is pulled in the third batch. Since the learner cannot update its policy inside the third batch, the vector $X_1, \dots, X_N$ denote the number of times that arm $i$ is pulled in the third batch. Since the learner cannot update its policy inside the third batch, the vector $(X_1, \dots, X_N)$ is determined by the observations available at the end of the second batch. Define the binary test
\[
\phi =
\begin{cases}
+, & X_u \ge X_v,\\
-, & X_u < X_v.
\end{cases}
\]
By the Bretagnolle-Huber inequality, we have 
\[
\Pr_+[\phi = -] + \Pr_-[\phi = +] \geq \frac{1}{2}e^{-4\delta^2 (q_u + q_v)}.
\]
Under such false events where suboptimal arms that have a gap of at least $\delta$ have been pulled at least $b_3/2$ times, the regret is at least $\frac{N^3 \delta}{2}$. Let $\pi_+ = \Pr[S=+ \mid E,U=u, V=v]$ and $\pi_- = \Pr[S=- \mid E,U=u, V=v]$. Since $\pi_+,\pi_- \geq c_0$, we have
\begin{align*}
\mathbb{E}[R_3 | F, U=u, V=v] &\geq \frac{N^3 \delta}{2} \cdot \left(\pi_+ \Pr_+[\phi = -] + \pi_- \Pr_-[\phi = +]\right)\\ 
&\geq \frac{c_0 N^3 \delta}{4}\exp(-4\delta^2 (q_u + q_v)).
\end{align*}
Taking expectation over $(U, V)$ conditioned on $F$ and using Jensen's inequality gives
\begin{align*}
\mathbb{E}[R_3 | F] &\geq \frac{c_0N^3 \delta}{4} \mathbb{E}[\exp(-4\delta^2 (q_U + q_V))|F]\\
&\geq \frac{c_0N^3 \delta}{4} \exp(-4\delta^2 \mathbb{E}[(q_U + q_V)|F])
\end{align*}
Recall that $\mathbb{E}[q_U + q_V | F] = \mathcal{O}(N)$ and $\delta = \frac{1}{16\sqrt{N}}$. Therefore, $\delta^2 \mathbb{E}[(q_U + q_V)|F] = \mathcal{O}(1)$, and
\[
\mathbb{E}[R_3 | F] = \Omega\left(N^3 \frac{1}{\sqrt{N}}\right) = \Omega\left(N^{\frac{5}{2}}\right).
\]

Finally, recall that $\Pr[F] = \Omega(1)$, we have 
\[
\mathbb{E}[\mathrm{Reg}] \geq \Pr[F] \mathbb{E}[R_3 | F] = \Omega(N^{5/2}) = \Omega(\alpha^{\frac{3}{2}} \sum_{i:\Delta_i > 0} \Delta_i^{-1}),
\]
where the last step follows the fact that one special arm has a gap $\delta$ in every instance in support of the hard distribution, while the remaining $N-2$ non-special arms have gaps that are bounded by universal constants, which gives $\sum_{i:\Delta_i>0 \frac{1}{\Delta_i}} = \frac{1}{\delta} + \Theta(N) = \Theta(N)$.

\paragraph{Randomized Algorithms.}
Let one randomized algorithm $A$'s internal randomness be $w$. Fixing $w$ makes the randomized algorithm a deterministic algorithm, denoted as $A_w$, where we have shown that for any deterministic algorithm, there exists a worst instance from a hard prior $\Pi$ that makes $\mathbb{E}_{\nu \sim \Pi}[\mathrm{Reg}] = \Omega(N^{5/2})$. Therefore, 
\[
\mathbb{E}_w \mathbb{E}_{\nu} [\mathrm{Reg}] = \mathbb{E}_{\nu} \mathbb{E}_w [\mathrm{Reg}] = \Omega(N^{5/2}) = \Omega\left( \alpha^{\frac{3}{2}} \sum_{i: \Delta_i > 0} \frac{1}{\Delta_i} \right),
\]
which completes the proof.
\end{proof}

\subsection{An Abstract Counterexample of the Current Gap-Dependent Analysis}
\label{sec:stochastic_algorithm_lower}
We examine the comparison step used in the analysis of \Cref{algo:stochastic_bandits}. Specifically, we construct an abstract instance with sequences of batch lengths and active-set sizes for which this comparison requires a factor of $\Omega(\alpha^{7/4})$. 

\begin{proposition}
There exists a family of abstract sequences $(b_j)_{j=1}^B$
and non-increasing positive integers $(n_j)_{j=1}^{B+1}$
such that, writing
\[
S_j := \sum_{k=1}^j \frac{b_k}{n_k},\qquad
\mathrm{LHS}_j := (n_j-n_{j+1})\frac{S_j}{\sqrt{S_{j-1}}},
\;
\mathrm{RHS}_j := (n_j-n_{j+1})\sqrt{S_{j-1}}
\; (j=2,\ldots,B),
\]
we have
\[
\frac{\sum_{j=2}^B \mathrm{LHS}_j}
     {\sum_{j=2}^B \mathrm{RHS}_j}
= \Theta(\alpha^{7/4}),
\]
where $\alpha$ is the hardness metric of $(b_j)_{j=1}^B$.
\end{proposition}
\begin{proof}
Let $N$ range over eighth powers of integers, and consider a game of five batches with lengths $b_j = N^{\frac{j}{2}}$, i.e., 
$
b_1 = N^{\frac{1}{2}}, b_2 = N, b_3 = N^{\frac{3}{2}}, b_4 = N^2, b_5 = N^{\frac{5}{2}}. 
$
By definition, we have $\alpha = \max_j \frac{b_j}{t_{j-1}} = \Theta(N^{1/2})$. Choose the abstract active-set size as
\[
n_1 = n_2 = n_3 = N, n_4 = N^{\frac{3}{4}}, n_5 = N^{\frac{3}{8}}, n_6 = 1.
\]
For $j < 3$, it is easy to see that $\mathrm{LHS}_j = \mathrm{RHS}_j = 0$ as $n_j - n_{j+1} = 0$. Moreover, we have $S_2 = \frac{N^{1/2}}{N} + \frac{N}{N} = \Theta(N/N) = \Theta(1)$. Similarly, we have $S_3 = \Theta(N^{\frac{3}{2}}/N) = \Theta(N^{1/2})$, $S_4 = \Theta(N^{5/4})$, and $S_5 = \Theta(N^{17/8})$. Plugging the above into $\sum_j \mathrm{LHS}_j$ and $\sum_j \mathrm{RHS}_j$ gives 
\[
\sum_{j=1}^B \mathrm{LHS}_j = \Theta(1) \cdot \left( N^{3/2} + N^{7/4} + N^{15/8} \right) = \Theta\left( N^{15/8} \right)
\]
and 
\[
\sum_{j=1}^B \mathrm{RHS}_j = \Theta(1) \cdot \left( N + N + N \right) = \Theta(N).
\]
Therefore, we have $\sum_j \mathrm{LHS}_j / \sum_j \mathrm{RHS}_j = N^{15/8} / N = N^{7/8} = \alpha^{7/4}$. Thus, no uniform comparison of these two sums over such abstract sequences can have a multiplicative factor $o(\alpha^{7/4})$.
\end{proof}

\section{Online Convex Optimization}

Online Gradient Descent (OGD) is a classical algorithm for online convex optimization. In the adversarially batched setting, the same algorithm applies, with the only change being that the update is performed once per batch using the aggregate subgradient. Specifically, starting from an arbitrary $x^{(1)} \in \mathcal{K}$, at the end of batch $j$ the learner performs the update 
\[
x^{(j+1)} = \Pi_{\mathcal{K}} \left( x^{(j)} - \eta z_j \right),
\]
where $\Pi_{\mathcal{K}}(y) = \arg\min_{x \in \mathcal{K}} \|x-y\|_2$ is Euclidean projection onto $\mathcal{K}$, and $\eta$ is the learning rate. 

\subsection{Upper Bound.}
\label{app:oco_upper}
\begin{proof}[Proof of \Cref{thm:oco_expert_adversarial_upper} (Point 1)]
Fix $x^* \in \mathcal{K}$. For each batch $j$, we define the aggregate loss function $F_j(x) = \sum_{t=t_{j-1}+1}^{t_j} f_t(x)$. Recall that $z_j = \sum_{t=t_{j-1}+1}^{t_j} g_t$. As the sum of convex functions is convex, $z_j \in \partial F_j\left(x^{(j)}\right)$, which implies
\[
F_j\left( x^{(j)} \right) - F_j\left( x^* \right) \leq \left\langle z_j, x^{(j)} - x^* \right\rangle.
\]
Therefore,
\[
\sum_{j=1}^B \left( F_j\left( x^{(j)} \right) - F_j\left( x^* \right) \right) \leq \sum_{j=1}^B \left\langle z_j, x^{(j)} - x^* \right\rangle.
\]
It remains to bound the right-hand side using the standard projected OGD argument. By the non-expansiveness of Euclidean projection, and since $x^* \in \mathcal{K}$, $\Pi_{\mathcal{K}}(x^*) = x^*$, so
\[
\left\| x^{(j+1)} - x^* \right\|_2^2 = \left\| \Pi_{\mathcal{K}} \left( x^{(j)} - \eta z_j \right) - \Pi_{\mathcal{K}} \left( x^* \right) \right\|_2^2.
\]
Expanding the square gives
\begin{align*}
    \left\| x^{(j+1)} - x^* \right\|_2^2 &\leq \left\| x^{(j)} - x^* \right\|_2^2 - 2\eta \left\langle z_j, x^{(j)} - x^* \right\rangle + \eta^2 \|z_j\|_2^2\\
    \iff \left\langle z_j, x^{(j)} - x^* \right\rangle &\leq \frac{\left\| x^{(j)} - x^* \right\|_2^2 - \left\| x^{(j+1)} - x^* \right\|_2^2}{2\eta} + \frac{\eta}{2} \|z_j \|_2^2.
\end{align*}
Summing over $j = 1, \dots, B$ gives
\[
\sum_{j=1}^B \left\langle z_j, x^{(j)} - x^* \right\rangle \leq \frac{\left\| x^{(1)} - x^* \right\|_2^2 - \left\| x^{(B)} - x^* \right\|_2^2}{2\eta} + \frac{\eta}{2} \|z_j \|_2^2.
\]
As $\|x^{(1)} - x^* \|_2 \leq D$, we have 
\[
\sum_{j=1}^B \left\langle z_j, x^{(j)} - x^* \right\rangle \leq \frac{D^2}{2\eta} + \frac{\eta}{2} \sum_{j=1}^B \| z_j \|_2^2 \leq \frac{D^2}{2\eta} + \frac{\eta G^2}{2} \|b\|_2^2,
\]
which completes the proof.
\end{proof}


\subsection{Lower Bound}
\label{app:oco_lower}
We further show that the dependence on $D, G$ and $\| b \|_2$ is unavoidable. It suffices to consider linear losses, which form a subclass of convex losses. 

\begin{proof}[Proof of \Cref{thm:oco_expert_adversarial_lower} (Point 1)]
Choose $x^+ , x^- \in \mathcal{K}$ such that $\| x^+ - x^- \|_2 = D$ and define $u = \frac{x^+ - x^-}{D}$. For every batch $j$, independently sample a Rademacher random variable $Z_j \in \{ -1, +1 \}$ and let the environment use the linear loss
\[
f_t(x) = GZ_j \langle u, x \rangle
\]
throughout batch $j$. Since the learner chooses $x^{(j)}$ before observing $Z_j$, 
\[
\mathbb{E} \left[ G b_j Z_j \langle u, x^{(j)} \rangle | Z_1, \dots, Z_{j-1} \right] = 0.
\]
Hence the learner's expected cumulative loss is $0$. 

For the best fixed decision term, since the width of $\mathcal{K}$ in direction $u$ is $D$, $\max_{x \in \mathcal{K}} \langle u, x \rangle - \min_{x \in \mathcal{K}} \langle u, x \rangle = D$. By symmetry of $S$, 
\[
\mathbb{E} \left[ \min_{x \in \mathcal{K}} G \langle u, x \rangle \sum_{j=1}^B  b_j Z_j = - \frac{DG}{2} \mathbb{E}\left| \sum_{j=1}^B  b_j Z_j \right| \right],
\]
which indicates that 
\[
\mathbb{E}[\mathrm{Reg}] = 0 - \left( -\frac{DG}{2} \mathbb{E}\left| \sum_{j=1}^B  b_j Z_j \right| \right) = \frac{DG}{2} \mathbb{E}\left| \sum_{j=1}^B  b_j Z_j \right|.
\]
Applying the Khintchine inequality (\Cref{lemma:khintchine}) with $p=1$ gives a regret lower bound of $\Omega \left( DG \sqrt{\sum_{j=1}^B b_j^2} \right) = \Omega(DG \| b \|_2)$ and completes the proof. 
\end{proof}

\section{Experts Problem}
Hedge\footnote{The algorithm is also known as the Exponential Weighted Average (EWA).}~\citep{freund1997decision} is a classical algorithm for the experts problem. In the non-batched setting, the algorithm maintains a positive weight $w_i^{(t)}$ for each expert $i \in [N]$ at each round $t \in [T]$. The learner then samples an expert according to the probability distribution $p_i^{(t)} = \frac{w_i^{(t)}}{\sum_{j=1}^N w_j^{(t)}}$ induced by the weights. After observing the losses, the algorithm updates the weights by assigning smaller weights to experts that incur larger losses with a learning rate $\eta > 0$:
\[
    w_i^{(t+1)} = w_i^{(t)} \exp(-\eta \ell_i^{(t)}).
\]
In the adversarially batched setting, the same idea applies, where updates are performed using the cumulative loss observed for each batch. Note that the weights must remain constant within each batch, i.e., $w_i^{(t_{j-1}+1)} = \dots = w_i^{(t_j)}$. Thus, we re-define the notation $w_i^{(j)}$ as the weight for expert~$i$ at the beginning of batch~$j$ in the rest of the paper, and write $p_i^{(j)}$ for the associated probability accordingly. A pseudo-code is provided in \Cref{algo:batched_hedge}.

\begin{algorithm}
\caption{Hedge in the Adversarially Batched Experts Setting}
\label{algo:batched_hedge}
\begin{algorithmic}[1]
\State Initialize: $w_i^{(1)} \gets 1\ \forall i\in[N]$; $j \gets 1$
\For{$t = 1, \ldots, T$}
    \State $p_i^{(j)} \gets \dfrac{w_i^{(j)}}{\sum_{k=1}^N w_k^{(j)}}\ \forall i\in[N]$
    \State Sample expert $i_t \sim p^{(j)}$
    \If{cumulative losses are revealed to the player}
        \State $t_j \gets t$
        \State Observe cumulative loss $L_i^{(j)} \; \forall i\in[N]$
        \State Update weights:
        $
        w_i^{(j+1)} \gets w_i^{(j)} \exp(-\eta L_i^{(j)}) \qquad \forall i\in[N]
        $
        \State $j \gets j + 1$
    \EndIf
\EndFor
\end{algorithmic}
\end{algorithm}

\subsection{Upper Bound}
\label{sec:expert_upper}
We show that Hedge naturally adapts to the adversarially batched experts setting. 
The only change from the standard algorithm is that the multiplicative update is applied once per batch, using the cumulative loss
$
L_i^{(j)} \coloneqq \sum_{t=t_{j-1}+1}^{t_j} \ell_i^{(t)} \in [0,b_j].
$
Thus, within each batch, the distribution remains fixed, and after observing the cumulative batch losses, the weights are updated as
$
w_i^{(j+1)} = w_i^{(j)} \exp(-\eta L_i^{(j)}).
$
The second point of the proof of \Cref{thm:oco_expert_adversarial_upper} then follows from a simple modification to the standard Hedge analysis.

\begin{lemma}[Hoeffding's Lemma]
\label{lemma:hoeffding_lemma}
    Let $X$ be a real-valued random variable supported on an interval $[a,b]$. Then for any $\eta \in \mathbb{R}$,
    $$\ln\mathbb{E}\left[e^{-\eta X}\right]\leq -\eta \mathbb{E}\left[X\right] +{\frac{\eta^{2}\left(b-a\right)^{2}}{8}}.$$
\end{lemma}

\begin{proof}[Proof of \Cref{thm:oco_expert_adversarial_upper} (Point 2)]
    Let $Z_j := \sum_{i=1}^N w_i^{(j)}$ be the total weight of all $N$ experts at block $j$, which makes $p_i^{(j)} = \frac{w_i^{(j)}}{Z_j}$
    We conduct the proof by bounding $\ln Z_{B+1}$ and recover the inequality in the regret formula.
    Recall that in the Hedge algorithm, the weight of each expert $i$ is updated according to $w_i^{(j+1)} = w_i^{(j)} \cdot e^{-\eta L_i^{(j)}}$. It follows that $w_i^{(B + 1)} = \exp \left( -\eta \sum_{j=1}^B L_i^{(j)} \right)$, which can be used to lower bound $\ln Z_{B}$ as follows:
    \[
    \ln Z_{B+1} = \ln\left( \sum_{i=1}^N \exp \left( -\eta\sum_{j=1}^BL_i^{(j)}\right) \right)
    \geq \max_{i \in [N]}  -\eta\sum_{j=1}^BL_i^{(j)}
    =   -\eta \min_{i \in [N]}\sum_{j=1}^BL_i^{(j)}.
    \]
    
    For the lower bound, we first bound the ratio $\frac{Z_{j+1}}{Z_{j}}$ using \Cref{lemma:hoeffding_lemma}. As $p_i^{(j)} = \frac{w_i^{(j)}}{Z_j}$, we have
    \[
        \ln \frac{Z_{j+1}}{Z_{j}} = \ln \sum_{i=1}^N\left(\frac{w_i^{(j)}}{Z_{j}}\exp\left(-\eta L_i^{(j)}\right)\right)
        = \ln \mathbb{E}_{i} \left[\exp\left(-\eta L_i^{(j)}\right) \right].
    \]
    Applying \Cref{lemma:hoeffding_lemma} to $L_i^{(j)} = \sum_{k=t_{j-1}+1}^{t_{j}}l_i^{(k)} \in [0, b_j]$ and summing over all batches gives
    \[
        \sum_{j=1}^B\ln \frac{Z_{j+1}}{Z_{j}}
    \le \sum_{j=1}^B \left(-\eta \cdot \mathbb{E}\left[L^{(j)}\right] + \frac{\eta^2b_j^2}{8} \right)
    =   -\eta \sum_{j=1}^B \mathbb{E}\left[L^{(j)}\right] + \frac{\eta^2}{8}\sum_{j=1}^B b_j^2.
    \]
    Since the left-hand side above is equal to $\ln Z_{B + 1} - \ln Z_1 = \ln Z_{B + 1} - \ln N$ by telescoping,
    \[
        \ln Z_{B+1} \leq \ln N -\eta \sum_{j=1}^B \mathbb{E}\left[ L^{(j)} \right] + \frac{\eta^2}{8}\sum_{j=1}^B b_j^2.
    \]

    Combining the above with the lower bound gives
    \[
        -\eta \min_{i^* \in [N]}\sum_{j=1}^BL_i^{(j)}
    \le \ln N -\eta \sum_{j=1}^B \mathbb{E}[L^{(j)}] + \frac{\eta^2}{8}\sum_{j=1}^B b_j^2,
    \]
    which implies the regret bound
    \[
        \sum_{j=1}^B\Ex{}{L^{(j)}} - \min_{i^* \in [N]}\sum_{j=1}^BL_i^{(j)}
    \le \frac{\ln N}{\eta} + \frac{\eta}{8}\sum_{j=1}^B b_j^2.
    \]
    Setting $\eta = \sqrt{\frac{8\ln N}{\sum_{j=1}^B b_j^2}}$ gives a regret bound of $\calO(\|b\|_2\sqrt{\log N})$. This completes the proof, since the regret is always trivially bounded by $T$.
\end{proof}

\subsection{Lower Bound}
\label{sec:expert_lower}
Towards proving the second point of \Cref{thm:oco_expert_adversarial_lower}, we first prove a lower bound for any arbitrary batch structure, and then construct a broad family of hard instances that gives a matching worst-case lower bound. Specifically, we first prove \Cref{lemma:expert_2_lower_bound}, which gives an $\Omega(\|b\|_2)$ regret lower bound for the $N = 2$ case. Then, we prove the second point of \Cref{thm:oco_expert_adversarial_lower} by bootstrapping from \Cref{lemma:expert_2_lower_bound} via dividing the $N$-expert instance into $\approx \log_2 N$ independent two-expert instances.

\begin{lemma}
\label{lemma:expert_2_lower_bound}
    For the adversarially batched expert problem with $N=2$ experts and batch structure $\calB = (b_1, b_2, \ldots, b_B)$, the minimax regret is at least $\Omega(\|b\|_2)$.
\end{lemma}


We prove \Cref{lemma:expert_2_lower_bound} using a standard construction, in which the losses are from independent coin flips, and the Khintchine inequality (\Cref{lemma:khintchine}). 

\begin{lemma}[Khintchine inequality]
\label{lemma:khintchine}
Let $\{Z_i\}_{i=1}^N$ be i.i.d. random variables with $\pr{}{Z_i = +1} = \pr{}{Z_i = -1} = 1/2$ for each $i \in [N]$. Let $0 < p < \infty$ and let $x_1, \ldots, x_i \in \mathbb{C}$. Then 
$$A_p\left(\sum_{i=1}^N |x_i|^2\right)^{\frac{1}{2}} \leq \left(\Ex{}{\left|\sum_{i=1}^N x_iZ_i\right|^p}\right)^\frac{1}{p} \leq B_p \left(\sum_{i=1}^N |x_i|^2\right)^{\frac{1}{2}}$$
for some constants $A_p, B_p > 0$. 
\end{lemma}

\begin{proof}[Proof of \Cref{lemma:expert_2_lower_bound}]
    Consider two experts who play opposite actions at random within each batch. Let $X_1, X_2, \ldots, X_B$ be independent random variables that follow $\Bern(1/2)$. The losses are specified by $L^{(j)}_1 = b_j \cdot X_j$ and $L^{(j)}_2 = b_j \cdot (1 - X_j)$ for each $j \in [B]$, and we write $L^{(j)} = \left(L^{(j)}_1, L^{(j)}_2\right)$ as the (two-dimensional) loss vector. Over the randomness in the losses (i.e., in $X$), the expectation of the regret is given by
    \[
        \Ex{X}{\sum_{j=1}^B \left\langle p^{(j)}, L^{(j)} \right\rangle} - \Ex{X}{\min_{i^* \in [N]} \sum_{j=1}^B L_{i^*}^{(j)}}.
    \]

    For the first term, note that $p^{(j)}$ only depends on $X_1, X_2, \ldots, X_{j-1}$. Thus, $X_j$ still follows $\Bern(1/2)$ conditioned on $p^{(j)}$. It follows that $\Ex{X}{\left\langle p^{(j)}, L^{(j)} \right\rangle} = b_j / 2$. For the second term, we have
    \[
        \Ex{X}{\min_{i^* \in [N]} \sum_{j=1}^B L_{i^*}^{(j)}}
    =   \Ex{X}{\min\left\{\sum_{j=1}^{B}b_j \cdot X_j, \sum_{j=1}^{B}b_j \cdot (1 - X_j)\right\}}.
    \]
    Therefore, we can re-write the expected regret as
    
    \begin{align*}
            &~\sum_{j=1}^B \frac{b_j}{2} - \Ex{X}{\min\left\{\sum_{j=1}^{B}b_j \cdot X_j, \sum_{j=1}^{B}b_j \cdot (1 - X_j)\right\}}\\
        =   &~\Ex{X}{\max\left\{\sum_{j=1}^{B}b_j \cdot (1/2 - X_j), \sum_{j=1}^{B}b_j \cdot (X_j - 1/2)\right\}}\\
        =   &~\Ex{X}{\left|\sum_{j=1}^{B}b_j \cdot (X_j - 1/2)\right|}
        =   \frac{1}{2}\Ex{Z}{\left|\sum_{j=1}^B b_j Z_j \right|},
    \end{align*}
    where $Z_1, Z_2, \ldots, Z_B$ denote independent Rademacher random variables (i.e., $\pr{}{Z_j = +1} = \pr{}{Z_j = -1} = 1/2$) in the last step. Applying \Cref{lemma:khintchine} with $p = 1$ gives a regret lower bound of $\frac{A_1}{2} \cdot \sqrt{\sum_{j=1}^{B}b_j^2} = \Omega(\|b\|_2)$ and completes the proof.
\end{proof}

\Cref{lemma:expert_2_lower_bound} gives the lower bound of the adversarially batched expert problem under the $N = 2$ case. To handle the general-$N$ case, the intuition is to partition the $B$ blocks into $\approx \log_2 N$ independent \emph{stages}, such that the block sequence of each stage has a $2$-norm of $\approx \|b\|_2 / \sqrt{\log N}$. For each stage, we design only two different losses, so each stage can be viewed as an independent $2$-expert problem. We then apply \Cref{lemma:expert_2_lower_bound} to lower bound the regret within each stage, and argue that these regrets are additive. We start by proving the lemma below, which states that under the worst-case construction that satisfies $\max_{j \in [B]} b_j \le \calO(\|b\|_2 / \sqrt{\log N})$, we can partition the game into $\Omega(\log N)$ stages. This, in turn, gives an $\Omega(\|b\|_2\sqrt{\log N})$ worst-case lower bound.

\begin{lemma}
\label{lemma:balanced_partition_l2}
Suppose that $b_1, b_2, \ldots, b_{B} > 0$ and $\max_{j \in [B]} b_j \le c \cdot \frac{\|b\|_2}{\sqrt{\log_2 N}}$ for some $c > 0$. Then, there exists $S \ge \left\lfloor \frac{\log_2 N}{1 + c^2} \right\rfloor$ and a partition $\{G_1, G_2, \dots, G_S\}$ of $[B]$ such that, for each $s \in [S]$,
\[
    \sqrt{\sum_{j \in G_s} b_j^2}
\ge
    \frac{\|b\|_2}{\sqrt{\log_2 N}}.
\]
Moreover, each partition $G_s$ consists of contiguous integers, i.e., $G_s = \{j_0, j_0 + 1, \ldots, j_1\}$, for some integers $j_0 \le j_1$.
\end{lemma}

\begin{proof}
We construct the partition by greedily forming minimal segments, so that the sum of squared entries within each segment is $\ge M \coloneqq \|b\|_2^2 / (\log_2 N)$. Formally, we start with $t_0 = 0$. For each $s = 1, 2, \ldots$, we define $t_s$ to be the smallest index that satisfies
\[
    \sum_{j=t_{s-1}+1}^{t_s} b_j^2 \geq M.
\]
If such an index exists, we must have $\sum_{j=t_{s-1}+1}^{t_s - 1} b_j^2 < M$. If no valid $t_s$ exists, we set $S = s - 1$ and reset $t_S$ to $B$. The resulting partition is then given by $G_s = \{t_{s-1} + 1, t_{s-1}+2, \ldots, t_s\}$ for each $s \in [S]$. By construction, each $s \in [S]$ satisfies the desired inequality
\[
    \sqrt{\sum_{j \in G_s} b_j^2}
\ge \sqrt{M}
=   \frac{\|b\|_2}{\sqrt{\log_2 N}}.
\]

It remains to lower bound $S$, the number of segments. Note that for each $s \in [S - 1]$, we have
\[
    \sum_{j \in G_s}b_j^2
=   \sum_{j=t_{s-1}+1}^{t_s - 1}b_j^2 + b_{t_s}^2
\le M + \left(c \cdot \frac{\|b\|_2}{\sqrt{\log_2 N}}\right)^2
= (1 + c^2) \cdot \frac{\|b\|_2^2}{\log_2 N}.
\]
Furthermore, we must have $\sum_{j \in G_S}b_j^2 < 2 \cdot (1 + c^2) \cdot \frac{\|b\|_2^2}{\log_2 N}$; otherwise, we should have formed at least two segments from blocks $t_{S-1} + 1$~through~$B$. Therefore, we have
\[
    \|b\|_2^2
=   \sum_{s=1}^{S}\sum_{j \in G_s}b_j^2
<   (S + 1) \cdot (1 + c^2) \cdot \frac{\|b\|_2^2}{\log_2 N}.
\]
This gives $S > \frac{\log_2 N}{1 + c^2} - 1$ and thus $S \ge \left\lfloor\frac{\log_2 N}{1 + c^2}\right\rfloor$.
\end{proof}

Now we are ready to prove the second point of \Cref{thm:oco_expert_adversarial_lower}.
\begin{proof}[Proof of \Cref{thm:oco_expert_adversarial_lower} (Point 2)]
We apply \Cref{lemma:balanced_partition_l2} to partition the block representation $\calB = (b_1, \dots, b_{B})$ of the family of hard instances that satisfies $\max_{j \in [B]} b_j \leq \mathcal{O}(\|b\|_2 / \sqrt{\log N})$ into $S \ge \Omega(\log N)$ contiguous stages $G_1, G_2, \ldots, G_S$. Each $G_s = \{t_{s-1} + 1, t_{s-1} + 2, \ldots, t_s\}$ denotes the indices of the blocks within stage~$s$, and the number of time steps in stage~$s$ is $T_s = \sum_{j \in G_s} b_j$.

For each stage $s \in [S]$, we independently apply \Cref{lemma:expert_2_lower_bound} to construct two random loss sequences
\[
    \left(\ell^{(t_{s-1}+1)}_0, \ell^{(t_{s-1}+2)}_0, \ldots, \ell^{(t_s)}_0\right)
\quad \text{and} \quad
    \left(\ell^{(t_{s-1}+1)}_1, \ell^{(t_{s-1}+2)}_1, \ldots, \ell^{(t_s)}_1\right),
\]
where $\ell^{(j)}_i$ denotes the total loss of expert~$i \in \zo$ within batch $j \in G_s$.

Without loss of generality, we assume that there are exactly $N = 2^S$ experts.  For clarity, we label the experts using length-$S$ binary strings in $\zo^S$, i.e., each expert is labeled with $S$ bits $i_1, i_2, \ldots, i_S$. Then, the losses in the original instance are defined as
\[
    L^{(j)}_{i_1, i_2, \ldots, i_S} \coloneqq \ell^{(j)}_{i_s},
\]
where $s \in [S]$ is the unique index such that $j \in G_s$. In other words, the expert labeled $i = (i_1, i_2, \ldots, i_S)$ incurs the same loss as ``expert~$i_s$'' does within stage~$s$.

Now, suppose that the learner incurs a loss of $\Lhat^{(j)}$ within each block $j \in [B]$. Then, the regret of the learner is given by
\begin{align*}
    \sum_{j=1}^{B}\Lhat^{(j)} - \min_{i \in \zo^S}\sum_{j=1}^{B}L^{(j)}_i
&=  \sum_{s=1}^{S}\sum_{j \in G_s}\Lhat^{(j)} - \sum_{s=1}^{S}\min_{i_s \in \zo}\sum_{j \in G_s}\ell^{(j)}_{i_s}\\
&=  \sum_{s=1}^{S}\left[\sum_{j \in G_s}\Lhat^{(j)} - \min_{i_s \in \zo}\sum_{j \in G_s}\ell^{(j)}_{i_s}\right].
\end{align*}

Note that $\sum_{j \in G_s}\Lhat^{(j)} - \min_{i_s \in \zo}\sum_{j \in G_s}\ell^{(j)}_{i_s}$ is exactly the learner's regret within stage~$s$. Since we constructed the losses using \Cref{lemma:expert_2_lower_bound}, the regret within each stage~$s$ is at least $\Omega\left(\sqrt{\sum_{j \in G_s}b_j^2}\right)$, which is in turn lower bounded by $\Omega\left(\frac{\|b\|_2}{\sqrt{\log N}}\right)$ by our construction of the partition from \Cref{lemma:balanced_partition_l2}. Finally, since \Cref{lemma:balanced_partition_l2} guarantees $S \ge \Omega(\log N)$, the regret is lower bounded by $\Omega(\|b\|_2\sqrt{\log N})$, which marks the end of the proof.
\end{proof}




\section{Adversarial Bandits}
\label{app:adversarial_bandit}
Recall that in the adversarial bandits setting, the player only observes the loss of the chosen arm~$l_{i_t}^{(t)}$ at each round $t \in [T]$. The EXP3 algorithm of Auer, Cesa-Bianchi, Freund, and Schapire~\citep{auer2002nonstochastic} uses $\tilde{l}^{(t)} = \left(0, \ldots, {l_{i_t}^{(t)}}/{p_{i_t}^{(t)}}, \ldots, 0 \right)$ as an estimator for the actual loss vector $l^{(t)}$. This estimator is unbiased since, for each coordinate $i \in [N]$, 
\[
    \Ex{i_t \sim p^{(t)}}{\tilde{l}^{(t)}_i} = \pr{i_t \sim p^{(t)}}{i_t = i} \cdot \frac{l^{(t)}_{i_t}}{p_{i_t}^{(t)}} + \pr{i_t \sim p^{(t)}}{i_t \ne i} \cdot 0 = l^{(t)}_i.
\]

In the adversarially batched setting, we update the weights according to
\[
w_i^{(j+1)} = w_i^{(j)} \exp(-\eta \tilde{L}_i^{(j)}),
\]
where $\tilde{L}_i^{(j)} = \sum_{t=t_{j-1}+1}^{t_{j}} \tilde{l}_i^{(t)}$. The weights then determine the probability of pulling each arm via $p_i^{(j+1)} = \frac{w_i^{(j+1)}}{\sum_{n \in [N]} w_n^{(j+1)}}$. A detailed pseudo-code is provided in \Cref{algo:batched_exp3}.
\begin{algorithm}
\caption{EXP3 in the Adversarially Batched Bandits Setting}
\label{algo:batched_exp3}
\begin{algorithmic}[1]
\State Initialize: $w_i^{(1)} \gets 1\ \forall i\in[N]$, $p^{(1)} \gets \frac{w_i^{(1)}}{\sum_{k=1}^N w_k^{(1)}} \forall i \in [N]$, and $j \gets 1$
\State Sample arm~$i_j \sim p^{(j)}$
\For{$t = 1, \ldots, T$}
    \State $p_i^{(j)} \gets \dfrac{w_i^{(j)}}{\sum_{k=1}^N w_k^{(j)}} \; \forall i\in[N]$
    \If{bandit feedback for the current block is revealed to the player}
        \State $t_j \gets t$
        \State Observe block loss of the played arm:
        $
        L_{i_j}^{(j)}
        $
        \State Form importance-weighted loss estimates:
        $\widehat{L}^{(j)}_i \leftarrow \frac{L^{(j)}_{i}}{p^{(j)}_{i}}\, \mathbf{1}\{i=i_j\}$
        \State Update weights:
        $
        w_i^{(j+1)} \gets w_i^{(j)} \exp\!\left(-\eta \widehat{L}_i^{(j)}\right)
        \qquad \forall i\in[N]
        $
        \State $j \gets j + 1$
        \State Sample arm~$i_j \sim p^{(j)}$
    \EndIf
\EndFor
\end{algorithmic}
\end{algorithm}


\subsection{Upper Bound}
\label{app:adversarial_bandit_upper}
\begin{proof}[Proof of \Cref{thm:oco_expert_adversarial_upper} (Point 3)]
Let $Z_j := \sum_{i=1}^N w_i^{(j)}$ to be the total weight of all $N$ experts at block $j$, and then bound $\ln Z_{B+1}$ on both directions.

Fix an arm~$i^* \in [N]$. To lower bound $\ln Z_{B+1}$, we note
\[
    \ln Z_{B+1}
=   \ln\left(\sum_{i=1}^N \exp\left(-\eta \sum_{j=1}^B \tilde{L}_i^{(j)}\right)\right)
\ge -\eta \cdot \sum_{j=1}^B \tilde{L}_{i^*}^{(j)}.
\]
For the upper bound, recall that $p_i^{(j)} = \frac{w_i^{(j)}}{\sum_{n=1}^N w_n^{(j)}}$. Then,
\[
\frac{Z_{j+1}}{Z_j} = \sum_{i=1}^N \frac{w_i^{(j)}}{Z_j} \exp\left(-\eta \tilde{L}_i^{(j)}\right) = \sum_{i=1}^N p_i^{(j)} \exp\left(-\eta \tilde{L}_i^{(j)}\right).
\]
Applying the inequality $e^{-x} \le 1 - x + \frac{x^2}{2}$ for $x \ge 0$, we obtain 
\begin{align*}
    \frac{Z_{j+1}}{Z_j} &\leq \sum_{i=1}^N p_i^{(j)}\left[1 - \eta \tilde{L}_i^{(j)} + \frac{\eta^2}{2}\left(\tilde{L}_i^{(j)}\right)^2\right]\\
    &= 1 - \eta \sum_{i=1}^N p_i^{(j)}\tilde{L}_i^{(j)} + \frac{\eta^2}{2} \sum_{i=1}^N p_i^{(j)}\left(\tilde{L}_i^{(j)}\right)^2.
\end{align*}
Taking the logarithm and summing over all blocks leads to
\[
\ln Z_{B+1} - \ln Z_1
\leq \sum_{j=1}^B \ln \left(1 - \eta \sum_{i=1}^N p_i^{(j)} \tilde{L}_i^{(j)} + \frac{\eta^2}{2} \sum_{i=1}^N p_i^{(j)} \left( \tilde{L}_i^{(j)} \right)^2 \right). 
\]
Applying $\ln(1+x) \le x$ and noting $Z_1 = N$ gives
\[
    \ln Z_{B+1}
\le \ln N -\eta \sum_{j=1}^B \sum_{i=1}^N p_i^{(j)} \tilde{L}_i^{(j)} + \frac{\eta^2}{2} \sum_{j=1}^B \sum_{i=1}^N p_i^{(j)} \left( \tilde{L}_i^{(j)} \right)^2.
\]
Combining the above with the lower bound $\ln Z_{B+1} \ge -\eta \cdot \sum_{j=1}^B \tilde{L}_{i^*}^{(j)}$ gives
\[
    \eta \left( \sum_{j=1}^B \sum_{i=1}^N p_i^{(j)} \tilde{L}_i^{(j)} - \sum_{j=1}^B \tilde{L}_{i^*}^{(j)} \right)
\le \ln N + \frac{\eta^2}{2} \sum_{j=1}^B \sum_{i=1}^N p_i^{(j)} \left( \tilde{L}_i^{(j)} \right)^2.
\]
Equivalently,
\[
    \sum_{j=1}^B \sum_{i=1}^N p_i^{(j)} \tilde{L}_i^{(j)} - \sum_{j=1}^B \tilde{L}_{i^*}^{(j)}
\le \frac{\ln N}{\eta} + \frac{\eta}{2} \sum_{j=1}^B \sum_{i=1}^N p_i^{(j)} \left( \tilde{L}_i^{(j)} \right)^2.
\]

Recall that $p^{(j)}$ is itself a random variable and directly affects $\tilde{L}^{(j)}$. Given the definition of pseudo-regret, we check the expectation of $\sum_{j=1}^B \sum_{i=1}^N p_i^{(j)} \tilde{L}_i^{(j)} - \sum_{j=1}^B \tilde{L}_{i^*}^{(j)}$. Specifically,
\begin{align*}
    \Ex{p^{(j)}}{\sum_{j=1}^B \sum_{i=1}^N p_i^{(j)} \tilde{L}_i^{(j)} - \sum_{j=1}^B \tilde{L}_{i^*}^{(j)}}
&=  \sum_{j=1}^B \sum_{i=1}^N p_i^{(j)} \Ex{p^{(j)}}{\tilde{L}_i^{(j)}} - \sum_{j=1}^B \Ex{p^{(j)}}{\tilde{L}_{i^*}^{(j)}}\\
&=  \sum_{j=1}^B \sum_{i=1}^N p_i^{(j)} L_i^{(j)} - \sum_{j=1}^B  L_{i^*}^{(j)}.
\end{align*}
Taking the result back to the LHS of the inequality leads to 
\begin{align*}
    \sum_{j=1}^B \sum_{i=1}^N p_i^{(j)} L_i^{(j)} - \sum_{j=1}^B  L_{i^*}^{(j)}
&\le\Ex{p^{(j)}}{\frac{\ln N}{\eta} + \frac{\eta}{2} \sum_{j=1}^B \sum_{i=1}^N p_i^{(j)} \left( \tilde{L}_i^{(j)} \right)^2}\\
&= \frac{\ln N}{\eta} + \frac{\eta}{2} \sum_{j=1}^B \sum_{i=1}^N p_i^{(j)} \cdot \Ex{p^{(j)}}{\left( \tilde{L}_i^{(j)} \right)^2}\\
&= \frac{\ln N}{\eta} + \frac{\eta}{2} \sum_{j=1}^B \sum_{i=1}^N p_i^{(j)} \cdot p_i^{(j)} \cdot \left(\frac{L_i^{(j)}}{p_i^{(j)}}\right)^2\\
&= \frac{\ln N}{\eta} + \frac{\eta}{2} \sum_{j=1}^B \sum_{i=1}^N \left( L_i^{(j)} \right)^2\\
&\leq \frac{\ln N}{\eta} + \frac{\eta N}{2} \sum_{j=1}^B b_j^2. \tag{$L_i^{(j)} \in [0, b_i]$}
\end{align*}
Setting $\eta = \frac{\sqrt{2 \ln N}}{\|b\|_2\sqrt{N}}$ bounds the pseudo-regret by $\|b\|_2 \cdot \sqrt{2N \ln N} = \calO(\|b\|_2 \cdot \sqrt{N \log N})$.
\end{proof}

\subsection{Lower Bound}
Similar to the second point of the proof of \Cref{thm:oco_expert_adversarial_lower}, we first prove a lower bound for any arbitrary batch structure $\mathcal{B}$, then construct a family of hard instances that gives a tight worst-case lower bound.
\label{app:adversarial_bandit_lower}
\begin{proof}[Proof of \Cref{thm:oco_expert_adversarial_lower} (Point 3)]
Recall the notation that $b_{\max} = \max_{j \in [B]} b_j$. 

Let $K = \min \left\{ N, \max \left\{ 2, \lfloor \|b\|_2^2 / b_{\max}^2 \rfloor \right\} \right\}$. As $\| b \|_2^2 \geq b_{\max}^2$ and $N \geq 2$, we have 
\[
K \geq 2;\; K \leq 2 \|b\|_2^2 / b_{\max}^2.
\]
We construct $K$ hard environments, indexed by $\theta \in [K]$. Fixing a sufficiently small absolute constant $\gamma = 1/16$, we define $\delta_j = \gamma \frac{\sqrt{K} b_j}{\|b\|_2}$ for $j \in [B]$. It is easy to see that 
\[
\delta_j \leq \gamma \frac{\sqrt{K} b_{\max}}{\|b\|_2} \leq \gamma \sqrt{2} < \frac{1}{4}.
\]
For every $\theta \in [K]$, we define an oblivious randomized environment $P_{\theta}$ as follows. For every batch $j \in [B]$ and arm $i \in [N]$, independently sample 
\[
X_{j,i} \sim
\begin{cases}
\operatorname{Bernoulli}\left(\frac{1}{2}-\delta_j\right), & i=\theta,\\[4pt]
\operatorname{Bernoulli}\left(\frac{1}{2}\right), & i\neq\theta.
\end{cases}
\] 
Within batch $j$, set the loss of arm $i$ to be constant across all $b_j$ rounds, i.e., $l_i^{(t)} = X_{j, i};\; t = t_{j-1}+1,\dots, t_j$. Thus the cumulative loss of arm $i$ in batch $j$ is $L_i^{(j)} = b_j X_{j,i}$. Since the environment is oblivious and the random variables associated with batch $j$ are independent of the history before batch $j$, under $P_{\theta}$ we have
\[
\mathbb{E}\left[ L_{i_j}^{(j)} \right] = \Pr_{\theta}[i_j = \theta] b_j (\frac{1}{2} - \delta_j) + \Pr_{\theta}[i_j \neq \theta] \frac{b_j}{2} =  \frac{b_j}{2} - b_j\delta_j \Pr_{\theta}[i_j = \theta],
\]
where, recall that $i_j$ is the arm selected by the learner in batch $j$. On the other hand,
\[
\mathbb{E}_{\theta} \left[ L_{\theta}^{(j)} \right] = b_j (\frac{1}{2} - \delta_j),
\]
while every arm $i \neq \theta$ has expected loss $b_j / 2$. Therefore, the pseudo-regret under $P_{\theta}$ is 
\begin{align*}
    \overline{\mathrm{Reg}}_{\theta} &= \sum_{j=1}^B \left( \frac{b_j}{2} - b_j\delta_j\Pr[i_j = \theta] \right) - \sum_{j=1}^B b_j \left( \frac{1}{2} - \delta_j \right)\\
    &= \sum_{j=1}^B b_j \delta_j \left( 1-\Pr_{\theta}[i_j = \theta] \right)\\
    &= \sum_{j=1}^B b_j \delta_j - \sum_{j=1}^B b_j\delta_j \Pr_{\theta}[i_j = \theta] 
\end{align*}
For the first term, we have 
\begin{align*}
\sum_{j=1}^B b_j \delta_j = \gamma \frac{\sqrt{K}}{\|b\|_2} \sum_{j=1}^B b_j^2 = \gamma \sqrt{K} \|b\|_2.
\end{align*}
Now we show that, for at least one $\theta \in [K]$, the learner selects arm $\theta$ incorrectly on a constant fraction of this weighted mass. Define the null environment $P_0$, under which $X_{j,i} \sim \mathrm{Bern}(1/2)$ independently for every $j \in [B]$ and $i \in [N]$. We compare $P_0$ and $P_{\theta}$, Since the learner uses the same decision rule under $P_0$ and $P_\theta$, the only difference between the two transcript distributions occurs in batches in which arm $\theta$ is selected. By the chain rule for KL divergence, 
\[
D_{\mathrm{KL}} (P_0 \| P_{\theta}) = \sum_{j=1}^B \Pr_0 \left[ i_j = \theta \right] \mathrm{kl}(1/2 \| 1/2 - \delta_j),
\]
where $\mathrm{kl}(p\|q)$ denotes the binary relative entropy. For $0 \leq \delta_j \leq 1/4$, 
\[
\mathrm{kl}(1/2 \| 1/2 - \delta) = -\frac{1}{2} \log(1 - 4\delta^2) \leq \frac{8}{3}\delta_j^2.
\]
Hence, $D_{\mathrm{KL}} (P_0 \| P_{\theta}) \leq \frac{8}{3} \sum_{j=1}^B \Pr_0 \left[ i_j = \theta \right] \delta_j^2$. By Pinsker's inequality, for every $j \in [B]$,
\[
\Pr_{\theta} \left[ i_j = \theta \right] \leq \Pr_0 \left[ i_j = \theta \right] + \sqrt{\frac{1}{2} D_{\mathrm{KL}}(P_0 \| P_{\theta})}.
\]
Taking this back, we have
\[
\overline{\mathrm{Reg}}_{\theta} \geq \gamma \sqrt{K} \|b\|_2 - \sum_{j=1}^B b_j \delta_j \Pr_0[i_j= \theta] - \gamma \sqrt{K}\|b\|_2 \sqrt{\frac{1}{2} D_{\mathrm{KL}}(P_0 \| P_{\theta})}.
\]
Now we average the above inequality over $\theta \in [K]$. As the learner selects at most one arm in each batch, $\sum_{\theta=1}^K \Pr[i_j = \theta] \leq 1$, thus 
\[
\frac{1}{K} \sum_{\theta=1}^K \sum_{j=1}^B b_j \delta_j \Pr[i_j = \theta] \leq \frac{\sum_{j=1}^B b_j \delta_j}{K} = \frac{\gamma \sqrt{K} \|b\|_2}{K}.
\]
For the last term, the Cauchy-Schwarz inequality gives
\begin{align*}
\frac{1}{K}\sum_{\theta=1}^K \sqrt{\frac{1}{2} D_{\mathrm{KL}}(P_0 \| P_{\theta})} &\leq \sqrt{\frac{1}{K}\sum_{\theta=1}^K \frac{1}{2} D_{\mathrm{KL}}(P_0 \| P_{\theta})}\\
&\leq \sqrt{\frac{1}{K}\sum_{\theta=1}^K \frac{1}{2} \sum_{j=1}^B \Pr_0 [i_j = \theta] \frac{8}{3} \delta_j^2}\\
&= \sqrt{\frac{4}{3K} \sum_{j=1}^B \delta_j^2 \sum_{\theta=1}^K  \Pr_0 [i_j = \theta]}\\
&\leq \sqrt{\frac{4}{3K} \sum_{j=1}^B \delta_j^2}.
\end{align*}
Substituting $\delta_j$, we have 
\[
\sqrt{\frac{4}{3K} \sum_{j=1}^B \delta_j^2} = \sqrt{\frac{4}{3K} \gamma^2 K \frac{\sum_{j=1}^B b_j^2}{\|b\|_2^2}} = \frac{2\gamma}{\sqrt{3}}.
\]
Combining together, 
\begin{align*}
\frac{1}{K} \sum_{\theta=1}^K \overline{\mathrm{Reg}}_{\theta} &\geq \frac{1}{K} \sum_{\theta=1}^K \gamma \sqrt{K} \|b\|_2 - \frac{\gamma \sqrt{K \|b\|_2}}{K} - \gamma \sqrt{K} \|b\|_2 \frac{2\gamma}{3}\\
&= \gamma \sqrt{K} \|b\|_2 \left(  1 - \frac{1}{K} - \frac{2\gamma}{\sqrt{3}} \right).
\end{align*}
As $K \geq 2$ and $\gamma = 1/16$, it is easy to see that $\frac{1}{K} \sum_{\theta=1}^K \overline{\mathrm{Reg}}_{\theta} = \Omega(\sqrt{K \|b\|_2^2})$. From the definition of $K$, 
\[
K \geq \frac{1}{2} \min \left\{ N, \frac{\|b\|_2^2}{b_{\max}^2} \right\}.
\]
Therefore, 
\[
\sqrt{K \|b\|_2^2} = \Omega\left( \sqrt{\|b\|_2^2 \min \{ N,  \|b\|_2^2 / b_{\max}^2\}} \right) = \Omega \left( \min \left\{ \|b\|_2 \sqrt{N}, \frac{\|b\|_2^2}{b_{\max}} \right\} \right).
\]
Finally, we consider a hard instance where $b_{\max} \leq \mathcal{O}(\|b\|_2 / \sqrt{N})$. Under the $\min$ operation above, this hard instance yields a worst-case lower bound of $\Omega\left( \| b \|_2 \sqrt{N} \right)$, which ends the proof.
\end{proof}

\section{Reduction from Scale-Free Online Learning}
\label{app:scale_free_reduction}
We perform the reduction by treating each batch as a single meta-round. In batch $j$, the learner uses the decision produced by the online algorithm throughout $I_j = \{ t_{j-1}+1, \dots, t_j \}$, and feeds the aggregate loss over the batch back to the algorithm at time $t_j$. 

\subsection{OCO and Experts}
\label{app:reduction_oco_expert}
We use the algorithm, namely, SOLO FTRL, and theorems proposed in \cite{orabona2018scale}.
\begin{lemma}[Corollary~1 of \cite{orabona2018scale}]
\label{lemma:solo_ftrl}
Let $\mathcal{K}$ be a nonempty bounded closed convex set equipped with a norm $\| \cdot \|$, and let $\psi: \mathcal{K} \to \mathbb{R}_{+}$ be lower semicontinuous and $1$-strongly convex with respect to $\| \cdot \|$. Define $S = \sup_{x \in \mathcal{K}} \psi(x)$. Then SOLO FTRL with regularizer 
\[
R(w) = \frac{\sqrt{2.75}}{\sqrt{\sup_{v \in \mathcal{K}} f(v)}} f(w)
\]
satisfies
\[
\mathrm{Reg} \leq 13.3 \sqrt{\sup_{v \in \mathcal{K}} f(v) \sum_{j=1}^B \| z_j \|_*^2},
\]
for any sequence of loss vector $z_1, \dots, z_B$, where $\| \cdot \|_*$ is the dual norm. 
\end{lemma}
\paragraph{OCO.}
Let $x^{(j)} \in \mathcal{K}$ denote the action selected at meta-round $j$ and played throughout batch $j$, i.e., $x^{(t)} = x^{(j)}$ for every $t \in I_j$. For each $t \in I_j$, recall $g_t \in \partial f_t\left(x^{(j)}\right)$ is a subgradient at the decision used in batch $j$, and $z_j = \sum_{t \in I_j} g_t$. For every $u \in \mathcal{K}$, convexity gives
\[
\sum_{t=1}^T \left(f_t\left(x^{(t)}\right) - f_t(u)\right) \leq \sum_{j=1}^B \langle z_j, x^{(j)} - u \rangle. 
\]
Thus, the batched OCO problem is reduced to OLO on the meta-losses $z_j$. Applying \Cref{lemma:solo_ftrl} gives 
\[
\mathrm{Reg} \leq C \sqrt{\frac{D^2}{2} \sum_{j=1}^B \| z_j \|_2^2} = \mathcal{O} \left( DG \sqrt{\sum_{j=1}^B b_j^2} \right).
\]
\paragraph{Experts Problem.}
Let $l_t \in [0, 1]^N$ be the expert-loss vector at round $t$, and let $z_j = \sum_{t \in I_j} l_t$. Then $\| z_j \|_\infty \leq b_j$, and the regret in the batched experts problem is the OLO regret 
\[
\mathrm{Reg} = \sum_{j=1}^B \langle p_j, z_j \rangle - \min_{p \in \Delta_N} \sum_{j=1}^B \langle p, z_j \rangle.
\]
Equip $\Delta_N$ with the $l_1$-norm and take 
\[
f(p) = \sum_{i=1}^N p_i \log p_i + \log N.
\]
By Pinsker's inequality, $f$ is $1$-strongly convex with respect to $\| \cdot \|_1$, and $\sup_{p \in \Delta_N} f(p) = \log N$. Since the dual norm of $l_1$ is $l_\infty$, \Cref{lemma:solo_ftrl} gives 
\[
\mathrm{Reg} \leq C \sqrt{\log N \sum_{j=1}^B \| z_j \|_\infty^2} \leq \mathcal{O} \left( \|b\|_2 \sqrt{\log N} \right).
\]

\subsection{Adversarial Bandits}
\label{app:reduction_adversarial}
We use UMAB-NN proposed in \cite{chen2023improved}, as well as its theorem guarantees. 
\begin{lemma}[Theorem~1 of \cite{chen2023improved}]
\label{lemma:umab_nn}
Consider the oblivious adversarial bandit problem with $N$ arms and nonnegative loss vectors $z_1, \dots, z_B \in \mathbb{R}_+^N$. The UMAB-NN algorithm satisfies 
\[
\mathbb{E}[\mathrm{Reg}] = \tilde{O} \left( \sqrt{N \sum_{j=1}^B \| z_j \|_\infty^2} \right).
\]
\end{lemma}
At the beginning of batch $j$, the bandit algorithm samples an arm~$i_j$, which is played throughout $I_j$. At the end of the batch, the learner observes $z_j,{i_j}$, the aggregate loss of the selected arm. Since $l_t \in [0, 1]^N$, it is easy to see that $z_j \in \mathbb{R}_+^N$ and $\| z_j \|_\infty \leq b_j$. Applying \Cref{lemma:umab_nn} gives 
\[
\mathbb{E}[\mathrm{Reg}] = \tilde{O} \left( \sqrt{N} \| b \|_2 \right).
\]

\section{Reduction from Delayed-Feedback Models}
\label{app:delayed_feedback_reduction}
Similar to \Cref{app:scale_free_reduction}, we show that existing delayed-feedback results also recover the same order of upper bounds for the OCO and the experts problem. The reduction is conducted by viewing the feedback generated at each round of a batch as being delayed until the end of that batch. Specifically, for every $t \in I_j$, we view the feedback generated at round $t$ as being delayed until the end of batch $j$, and define $\tau_t = t_j - t$. We use the delayed FTRL-PROX analysis of \cite{joulani2016delay}. The following lemma is a specialization of their Theorem~4 to linear losses. 

\begin{lemma}[Specialization of Theorem~4 of \cite{joulani2016delay}]
\label{lemma:delayed_ftrl}
Let $\mathcal{K}$ be a nonempty closed convex set equipped with a norm $\|\cdot\|$, and let $\psi:\mathcal{K}\to\mathbb{R}_+$ be $1$-strongly convex with respect to $\|\cdot\|$. Define $S = \sup_{x\in\mathcal{K}} \psi(x)$. Consider delayed online linear optimization with loss vectors $g_1,\dots,g_T$ satisfying $\|g_t\|_* \leq G$. Let $\widetilde{\tau}_s$ denote the update delay of the $s$-th feedback processed by the SOLID reduction of \cite{joulani2016delay}. For any non-increasing sequence
$\eta_1,\dots,\eta_T>0$, FTRL-PROX with cumulative regularizer
$
r_{0:s}(x)=\frac{\psi(x)}{\eta_s}
$
satisfies
\[
\mathrm{Reg}
\leq
\frac{S}{\eta_T}
+
\frac{G^2}{2}\sum_{s=1}^T \eta_s\left(1+2\widetilde{\tau}_s\right).
\]
Moreover, $\sum_{s=1}^T \widetilde{\tau}_s = \sum_{t=1}^T \tau_t$.
\end{lemma}

Let $a_s = 1+2\widetilde{\tau}_s, A_s = \sum_{r=1}^s a_r$. Setting $\eta_s = \frac{\sqrt{S}}{G\sqrt{A_s}}$ is batch-oblivious, since $A_s$ depends only on feedback already received. Using $\sum_{s=1}^T a_s/\sqrt{A_s}\leq 2\sqrt{A_T}$, \Cref{lemma:delayed_ftrl} gives
$
\mathrm{Reg} \leq 2G\sqrt{S A_T}.
$
For our batched feedback pattern,
\[
\sum_{t=1}^T \tau_t = \sum_{j=1}^B \sum_{t=t_{j-1}+1}^{t_j}(t_j-t) = \frac{1}{2}\sum_{j=1}^B b_j(b_j-1).
\]
Therefore,
\[
A_T = T+2\sum_{t=1}^T\tau_t = \sum_{j=1}^B b_j^2 = \|b\|_2^2.
\]
\paragraph{OCO.}
Recall that $g_t\in\partial f_t(x^{(j)})$ for every $t\in I_j$, and let $ z_j=\sum_{t\in I_j}g_t$.  If only the aggregate first-order feedback $z_j$ is returned at the end of batch $j$, we may
equivalently feed the delayed algorithm $b_j$ copies of $ \bar g_j=\frac{z_j}{b_j}$,
all arriving at time $t_j$. Since $\|z_j\|_2\leq Gb_j$, we have $\|\bar g_j\|_2\leq G$.
Moreover, for every $u\in\mathcal{K}$, convexity gives
\[
\sum_{t=1}^T \left(f_t(x^{(t)})-f_t(u)\right) \leq \sum_{j=1}^B \langle z_j,x^{(j)}-u\rangle = \sum_{j=1}^B\sum_{t\in I_j} \langle \bar g_j,x^{(j)}-u\rangle.
\]
Thus, the batched OCO regret is upper bounded by the regret of the corresponding delayed OLO instance. Equip $\mathcal{K}$ with the Euclidean norm and take, for any fixed $x_0\in\mathcal{K}$,
\[
\psi(x)=\frac{1}{2}\|x-x_0\|_2^2.
\]
Then $\psi$ is $1$-strongly convex and $ S\leq \frac{D^2}{2}$. Applying \Cref{lemma:delayed_ftrl} and $A_T=\|b\|_2^2$ gives
\[
\mathrm{Reg} \leq 2G\sqrt{\frac{D^2}{2}\|b\|_2^2} = \mathcal{O}\left(DG\|b\|_2\right).
\]

\paragraph{Experts Problem.}
Let $z_j=\sum_{t\in I_j}l_t$ be the aggregate expert-loss vector in batch $j$. As above, at time $t_j$ we may feed the delayed algorithm $b_j$ copies of $\bar l_j=\frac{z_j}{b_j}$. Since $l_t\in[0,1]^N$, we have $\bar l_j\in[0,1]^N$ and
$\|\bar l_j\|_\infty\leq 1$. This transformation preserves the regret:
\[
\mathrm{Reg} = \sum_{j=1}^B\langle p_j,z_j\rangle - \min_{p\in\Delta_N}\sum_{j=1}^B\langle p,z_j\rangle = \sum_{j=1}^B\sum_{t\in I_j}\langle p_j,\bar l_j\rangle - \min_{p\in\Delta_N} \sum_{j=1}^B\sum_{t\in I_j}\langle p,\bar l_j\rangle.
\]
Equip $\Delta_N$ with the $l_1$-norm and take
\[
\psi(p)=\sum_{i=1}^N p_i\log p_i+\log N.
\]
By Pinsker's inequality, $\psi$ is $1$-strongly convex with respect to $\|\cdot\|_1$, and
\[
\sup_{p\in\Delta_N}\psi(p)=\log N.
\]
Since the dual norm of $l_1$ is $l_\infty$, applying \Cref{lemma:delayed_ftrl} with $G=1$ gives
\[
\mathrm{Reg}
\leq
2\sqrt{\log N}\,\|b\|_2
=
\mathcal{O}\left(\|b\|_2\sqrt{\log N}\right).
\]

\section{Naive Doubling Trick}
\label{sec:naive_doubling_proof}
As Intuitively shown in \Cref{tab:naive_comparison}, for each doubling epoch $r$, \Cref{algo:naive_doubling}  predicts the sum of squared known batch to be the actual sum of squared batch in epoch $r$, i.e., $\sum_{j=1}^{2^{r-1}} b_j^2 = \sum_{j=2^{r-1}+1}^{2^r} b_j^2$.

\begin{algorithm}
\caption{Naive Doubling Trick under Adversarial Batched Setting}
\label{algo:naive_doubling}
\begin{algorithmic}[1]
\State Play the uniform distribution until the first block ends (i.e., at time $b_1$);
\State $r \gets 1$; \Comment{start epoch~$1$}
\While{$2^r \leq B$}
    \State $g^{(r)} \gets \sum_{j=1}^{2^{r-1}} b_j^2$; \Comment{guess squared $2$-norm in epoch~$r$}
    \State $\eta^{(r)} \gets \sqrt{\frac{8 \ln N}{g^{(r)}}}$; \Comment{set learning rate}
    \State Run Hedge on blocks $2^{r-1}+1, \ldots, 2^r$ with learning rate $\eta^{(r)}$;
    \State $r \gets r+1$; \Comment{start a new epoch}
\EndWhile
\end{algorithmic}
\end{algorithm}

\begin{table}
\centering
\caption{Comparison of guessed and actual parameters across epochs.}
\label{tab:naive_comparison}
\begin{tabular}{c|c|c}
\hline
\textbf{Epoch ($r$)} & \textbf{Guessed ($g^{(r)}$)} & \textbf{Actual ($a^{(r)}$)} \\
\hline
1 & $b_1^2$ & $b_2^2$ \\
2 & $b_1^2 + b_2^2$ & $b_3^2 + b_4^2$ \\
3 & $b_1^2 + \ldots + b_4^2$ & $b_5^2 + \ldots + b_8^2$ \\
$\dots$ & $\dots$ & $\dots$\\
$R$ & $\sum_{j=1}^{2^{R-1}} b_j^2$ & $\sum_{j=2^{R-1}+1}^{2^R}b_j^2$\\
\hline
\end{tabular}
\end{table}

\begin{example}
\label{ex:doubling_trick}
Let $m \ge 2$ be an integer power of $2$.
Consider a batch structure with $B = 2m$ batches:
\[
    b_1 = \dots = b_m = 1,\; b_{m+1} = \dots = b_{2m} = m,
\]
and note that $\|b\|_2 = \sqrt{m^3 + m} = \Theta(m^{3/2})$.

Consider an instance with $N = 2$ experts. For simplicity, let both experts' per-round losses to 0 in small blocks ($b_1, \dots, b_m$) so that the first half of the game only serves to fool the naive algorithm. For the remaining $m$ blocks, the losses alternate between the two experts.
Specifically, for each $k=1,2,\ldots,m$, block $j=m+k$ satisfies
\[
\left(L_1^{(m+k)},\, L_2^{(m+k)}\right)
=
\begin{cases}
(0,\,m), & \text{if $k$ is odd}, \\[4pt]
(m,\,0), & \text{if $k$ is even}.
\end{cases}
\]
\end{example}

\begin{proof}[Proof of \Cref{propo:failure_doubling_trick}]
    Let $R$ be the index of the last epoch in \Cref{algo:naive_doubling}.
    We argue that by the construction of \Cref{ex:doubling_trick}, the actual number of blocks $a$ in epoch~$R$ is exactly the second half of the blocks, i.e., $a^{(R)} = \sum_{j=m+1}^{2m} m^2 = m^3$. On the other hand, \Cref{algo:naive_doubling} guesses the number of blocks in the last epoch to be $g^{(R)} = \sum_{j=1}^m 1^2 = m$, which largely deviates from $a^{(R)}$. The corresponding learning rate for epoch $R$ is then set to $\eta^{(R)} = \sqrt{\frac{8 \ln 2}{m}}$. In the following, we omit the superscript and write $\eta$ for the learning rate in the last epoch for simplicity.
    
    Since both actions have zero losses in the first $m$ blocks, we have $w_1^{(m+1)} = w_2^{(m+1)} = 1$ at the beginning of block~$m+1$. Since $\left(L_1^{(m+1)}, L_2^{(m+1)}\right) = (0,m)$, \Cref{algo:naive_doubling} incurs a loss of $\frac{m}{2}$, and updates the weights to $w_1^{(m+2)} = e^{-\eta \cdot 0} = 1$ and $w_2^{(m+2)} = e^{-\eta \cdot m} = e^{-\beta}$, where we use the shorthand $\beta \coloneqq m \cdot \eta = \sqrt{8m \ln 2}$.
    
    Then, in block $m+2$, the algorithm puts a much higher probability (of $p_1^{(m+2)} = \frac{1}{1 + e^{-\beta}}$) on expert~1, and incurs a loss of $\frac{m}{1 + e^{-\beta}}$. After that, the weights of the two experts are equal ($w_1^{(m+3)} = w_2^{(m+3)} = e^{-\beta}$), so the next block~$m + 3$ is identical to block~$m + 1$.
    
    Repeating this pattern, there are $\frac{m}{2}$ blocks (namely, blocks $m+1, m+3, \ldots$) on which \Cref{algo:naive_doubling} incurs a loss of $\frac{m}{2}$. On each of the remaining $\frac{m}{2}$ blocks, the loss is $\frac{m}{1+e^{-\beta}}$. Note that either experts has a total loss of $\frac{m}{2} \cdot m = m^2 / 2$ over the entire horizon. Therefore, the regret is given by
    \[
    \frac{m}{2} \cdot \frac{m}{2} + \frac{m}{2} \cdot \frac{m}{1+e^{-\beta}} - \frac{m^2}{2}
    = Cm^2
    = \Omega(m^2),
    \]
    where $C = \frac{1}{4} \cdot \frac{1 - e^{-\beta}}{1 + e^{-\beta}} = \frac{1}{4} \cdot \frac{1 - e^{-\sqrt{8m \ln 2}}}{1 + e^{-\sqrt{8m \ln 2}}} = \Omega(1)$ for all sufficiently large $m$.

    On the other hand, the desired regret bound from \Cref{thm:oco_expert_adversarial_upper} is
    \[
        \calO(\|b\|_2\sqrt{\log N})
    =   \calO(m^{3/2})
    \]
    on this instance. This completes the proof.
\end{proof}
\subsection{Adaptive Doubling Trick Sidesteps the Hard Instance}
We further instantiate \Cref{thm:adaptive_doubling} on the hard instance from \Cref{ex:doubling_trick}, showing that the adaptive doubling trick avoids the linear-regret failure mode of the naive doubling trick.

\begin{corollary}
\label{coro:adaptive_doubling_case}
For the adversarially batched experts instance constructed in \Cref{ex:doubling_trick}, the naive doubling-trick adaptation of Hedge incurs expected regret $\Omega(T)$, whereas the adaptive doubling trick described in \Cref{algo:doubling} achieves a regret of
$
\calO\left(T^{3/4}\right).
$
\end{corollary}
\begin{proof}
We consider the case of \Cref{ex:doubling_trick}, where \Cref{propo:failure_doubling_trick} has shown that the naive doubling trick has a regret of $\Omega (T)$. Applying \Cref{thm:adaptive_doubling}, we have
\[
\text{Reg}_N^{B} = \calO\left( \| b \|_2 \sqrt{ \log N }\right) = \calO\left( \sqrt{m \cdot m^2 \log 2}\right) = \calO\left(m^{\frac{3}{2}}\right).
\]
Taking $T \asymp m^2$ back into the equation completes the proof.
\end{proof}


\end{document}